\documentclass[11pt]{article}

\usepackage[final]{acl}
\usepackage{multirow}
\usepackage{times}
\usepackage{amsmath, amsfonts, amssymb,amsthm}
\usepackage{latexsym}
\usepackage{graphicx}
\usepackage{subcaption}
\usepackage{xcolor}
\usepackage[table]{xcolor}
\usepackage{hhline}
\usepackage{hhline}
\usepackage{booktabs}
\usepackage{afterpage}
\usepackage{graphicx}
\usepackage{float}
\usepackage{inconsolata}
\usepackage{listings}
\usepackage{stfloats}
\usepackage[most]{tcolorbox}
\tcbuselibrary{listings}
\tcbset{
    colback=gray!10,
    colframe=gray!50,
    arc=0pt,
    boxsep=1pt,
    left=4pt,
    right=4pt,
    top=1pt,
    bottom=1pt,
    boxrule=0.3pt
}
\usepackage{xcolor}
\usepackage{enumitem}

\newtcolorbox{promptbox}[2][]{
    title={#2},
    colback=white,
    colframe=black,
    colbacktitle=white,
    coltitle=black,
    boxed title style={
        colframe=black,
        colback=white
    },
    #1
}
\newtcolorbox{addbox}[1][]{
    colback=white,
    colframe=black,
    colbacktitle=white,
    coltitle=black,
    boxed title style={
        colframe=black,
        colback=white
    },
    #1
}

\usepackage[T1]{fontenc}

\usepackage[utf8]{inputenc}

\usepackage{microtype}

\usepackage{graphicx}

\title{Training dynamics of LLM synthetic data for structured tasks}
\title{Synthetic Data Characterization via Training Dynamics}

\author{First Author \\
  Affiliation / Address line 1 \\
  Affiliation / Address line 2 \\
  Affiliation / Address line 3 \\
  \texttt{email@domain} \\\And
  Second Author \\
  Affiliation / Address line 1 \\
  Affiliation / Address line 2 \\
  Affiliation / Address line 3 \\
  \texttt{email@domain} \\}

\author{Irene Lago,\textsuperscript{1} Ana Ezquerro,\textsuperscript{2} David Vilares\textsuperscript{1} \\
           \textsuperscript{1} Universidade da Coruña, CITIC, Spain\\
           \textsuperscript{2} Graz University of Technology, IML, Austria\\{\texttt{i.lago@udc.es}, \texttt{ana.ezquerro@tugraz.at}, \texttt{david.vilares@udc.es}}\\}

\usepackage{url}
\usepackage{scalerel}

\usepackage{xspace}
\newcommand{\llama}{$\mathcal{L}$}
\newcommand{\gemma}[1]{\ensuremath{\mathcal{G}_\text{#1}}}
\newcommand{\olmo}{$\mathcal{O}$}
\newcommand{\qwen}{$\mathcal{Q}$}

\newcommand{\wasserstein}{$\text{W}_1$\xspace}
\newcommand{\cliffdelta}{$\delta$\xspace}

\newcommand{\nan}{$\times$}

\begin{document}
\maketitle
\begin{abstract}
Interpreting properties of LLM-generated data is important for understanding its utility and limitations across learning tasks.
In this work, we characterize synthetic data through sample-level learnability, studying variation among LLM families and scales, alongside human-written data as a reference.
We first generate synthetic datasets spanning single- and multi-label classification, labeling, and tree prediction tasks. We then derive empirical data distributions from encoder training dynamics for both machine and organic data, and estimate the robustness of these distributions across encoders. Finally, we evaluate how data selection strategies based on these learnability signals affect both data sources differently.
\end{abstract}

\section{Introduction}

Synthetic data, comprising training or evaluation samples generated by large language models~\cite[LLMs;][]{radford2019language,brown-etal-2020-language,NEURIPS2022_b1efde53}, has become a key component of dataset construction for pre-training, instruction tuning, and fine-tuning~\cite{gunasekar2023textbooksneed, 
maini2024rephrasingwebrecipecompute, xu2025wizardlmempoweringlargepretrained}. Its importance has grown with rising data demands and the scarcity of new, high-quality human-written text~\cite{villalobos2024position}, 
motivating scalable annotation, broader domain coverage, controllable sample-level features, and lower-noise data.

Prior studies have shown the effectiveness of LLMs both as annotators~\cite{gilardi2023chatgpt} and as data generators for tasks including classification~\cite{wang-etal-2023-self-instruct,horych-etal-2025-promises}, grammatical error correction \cite{stahlberg-kumar-2024-synthetic}, coding~\cite{abed-etal-2025-increasing}, machine translation~\cite{wang-etal-2024-lambda}, and reasoning~\cite{ho-etal-2023-large,zelikman-etal-2022-star}.
However, synthetic text raises concerns regarding the diversity \cite{gude2026aligneddiverseanalyzinggrammar} and realism relative to human-written text, as it may deviate from the underlying natural data distribution~\cite{viswanathan-etal-2025-synthetic}.

In this context, recent work has explored strategies to assess the diversity and quality of synthetic data from linguistic perspectives~\cite{munoz2024contrasting,zamaraeva-etal-2025-comparing}, as well as frameworks that measure distributional or semantic similarity with respect to reference datasets across a range of downstream applications~\cite{li-etal-2023-synthetic,geng-etal-2025-alleviating,kim-etal-2025-evaluating}.
However, these approaches treat datasets as a static, monolithic distribution, rely on feature-space similarity metrics, and overlook the downstream behavior of individual samples during training.

\paragraph{Contribution} 
To address these limitations, we present a framework for characterizing synthetic data via sample-level learnability across diverse NLP tasks, including classification, sequence labeling, and tree-structured prediction; quantifying learnability using optimization dynamics and studying its behavior relative to reference benchmarks. First, we provide a comparative distributional analysis across LLMs and prompting strategies, measuring divergence between models and relative to human-written data.

Second, we assess the stability of these dynamics across encoders to evaluate robustness to backbone choice. Third, we evaluate difficulty-based data selection by training on difficulty strata and comparing against baselines.
Code is available in our public repository\footnote{\url{https://github.com/irenelago/SynthTD/}}.

\section{Background}

We review work on synthetic data generation, diversity and deviation from human text, and training-dynamics methods related to our framework.

\subsection{Synthetic data generation}\label{sec:synthetic-data-generation}
Producing synthetic data has long been challenging due to language ambiguity and contextual variability. Pre-LLM methods used operations to extend seed samples, like lexical perturbations \citep{wei-zou-2019-eda}, context-aware substitutions \citep{kobayashi-2018-contextual}, back-translation \citep{xie2020unsuperviseddataaugmentationconsistency}, and subtree edits \citep{dehouck-gomez-rodriguez-2020-data}. However, these methods often lead to low-diversity, near-duplicate examples.

LLMs overcome many of these historical issues \cite{wu-etal-2024-prompt}, producing higher-quality synthetic annotated data with minimal manual effort. While early work explored LLMs as substitutes for annotators on human-generated data \cite{gilardi2023chatgpt}, recent research increasingly focuses on autonomous data synthesis in zero-shot and few-shot settings, driven by the growing scarcity of high-quality natural text \cite{kang-etal-2025-demystifying}. For instance, prompting frameworks \cite{ye-etal-2022-zerogen, meng2022generatingtrainingdatalanguage} show that LLMs can synthesize high-quality training sets for text classification from scratch, using only class descriptors. This success extends to other domains such as question answering \cite{puri-etal-2020-training, harsha-etal-2025-synthetic} and information extraction \cite{dao-etal-2025-overcoming}. Beyond specific tasks, synthetic data has been used to elicit reasoning via Chain-of-Thought approaches \cite{zelikman-etal-2022-star}, showing that even small models can acquire such capabilities from synthetic explanation traces generated by teacher models \cite{mukherjee-etal-2023-orca}.

\subsection{Diagnosis of synthetic data}

Alongside their use for data generation, concerns have emerged regarding whether LLM-generated data reflects the underlying human data distribution or instead exhibits distributional biases.
Synthetic text is known to amplify biases \citep{santurkar2023opinionslanguagemodelsreflect, perez-etal-2023-discovering, hartmann2023politicalideologyconversationalai}, exhibit linguistic differences \cite{munoz2024contrasting,guo-etal-2024-curious,zamaraeva-etal-2025-comparing}, and can lead to model collapse \citep{shumailov2024, alemohammad-etal-2024-self-consuming}, where models progressively lose the ability to represent low-probability events.
To mitigate these issues, recent work proposes prompting techniques for generating diverse samples \citep{honovich2022unnaturalinstructionstuninglanguage, misaki2026stringseedthoughtprompting}, machine-generated text detectors for resampling toward more human-like outputs \citep{drayson-etal-2025-machine}, and verification methods for filtering erroneous data \citep{feng2024modelcollapsescalingsynthesized}.

These limitations motivate evaluations beyond surface-level quality \citep{gudibande-etal-2024-false}, e.g., focusing on learning impact. Data-driven approaches have measured similarity with respect to human data \cite{ramesh-etal-2025-synthtexteval, chim-etal-2025-evaluating} or a reference
dataset \citep{dankar}, and studied sample replicability \citep{el2024evaluation, alaa2022faithfulsyntheticdatasamplelevel}.

\subsection{Training dynamics}
Training dynamics \cite{wang-etal-2022-deep} studies sample-specific behavior during model optimization, with goals like improving  efficiency \cite{chimoto-etal-2024-critical, wang-etal-2026-predictive}, data pruning \cite{hanafi-etal-2025-identifying}, and understanding pretraining \cite{zhang-etal-2025-training, sun-etal-2025-theoretical} and finetuning curves \cite{wang-etal-2025-learning, du-etal-2025-ftft}. \citet{swayamdipta-etal-2020-dataset} observed these dynamics in classification NLP tasks on organic data by grouping instances based on learnability to improve performance. Notably, they introduced data maps—leveraging confidence, variability, and correctness metrics—to visualize and categorize examples.
\citet{mekala-etal-2024-smaller} modeled sample difficulty in LLM pretraining as function of learning speed, showing that small-model dynamics can be used to filter samples and reduce pretraining data for larger LLMs. \citet{sosea-caragea-2022-leveraging} adopted this concept for transfer learning, enhancing the pseudo-labeling in classification tasks to transfer knowledge from teacher to student models.

\paragraph{Training dynamics for synthetic data} 
The literature has begun to explore training-dynamics-based strategies for machine-generated data, e.g., by producing instances that match the embedding or gradient manifolds of models trained on human-authored data \cite{sucholutsky-etal-2021-soft, maekawa-etal-2024-dilm, nguyen-etal-2025-synthetic}. While these methods enhance generalization, they treat training dynamics as an optimization objective rather than a tool for diagnosing text. Bridging this gap, \citet{kang-etal-2025-demystifying} propose a framework to analyze the quality of synthetic data in LLM pretraining, finding that mixing rephrased and original text can accelerate convergence and that effective training mixtures depend on a complex trade-off between lexical and semantic diversity.

\section{Methodology}
We describe our approach to sample-level characterization of synthetic data via training dynamics. We first explain the metrics for sample difficulty and predictability derived from training behavior, then describe our synthetic data generation process.

\subsection{Data maps for classification and beyond}\label{sec:expanded-dataset-cartography}
\citet{swayamdipta-etal-2020-dataset} mapped training dynamics of human-authored instances to estimate their utility.
The framework targeted sequence-level classification, where samples $(\mathbf{x}, y) \sim \mathcal{D}$ are characterized by learning trajectories defined by \emph{confidence}, \emph{variability}, and \emph{correctness}.

Let  $\mathcal{V}$ denote a finite vocabulary of tokens, and $\mathbf{x}=(x_1\dots x_n)\in\mathcal{V}^n$ the input where each token is defined as  $x_i\in\mathcal{V}$, for $i\in\{1\dots n\}$. Let $\mathcal{Y}$ denote the target set, and let $p_{\theta_t}(y|\mathbf{x})$ represent the categorical distribution over $\mathcal{Y}$ conditioned on $\mathbf{x}$ estimated by a model with parameters $\theta$ at training step $t\in[1,T]$.  The confidence (Equation \ref{eq:confidence}) averages the expected accuracy of the model across training,  the variability (Equation \ref{eq:variability}) is the standard deviation of the confidence, and the correctness (Equation \ref{eq:correctness}) indicates how often the model produces the correct output on average.

\begin{equation}\label{eq:confidence}\small 
    \gamma(\mathbf{x},y) =\mathbb{E}_t[p_{\theta_t}(y|\mathbf{x})]=\frac{1}{T}\sum_{t=1}^T p_{\theta_t}(y|\mathbf{x})
\end{equation}
\begin{equation}\label{eq:variability}\small 
    \sigma(\mathbf{x},y) = \text{std}_t[p_{\theta_t}(y|\mathbf{x})] = \sqrt{\mathbb{E}_t\left[\left(p_{\theta_t}(y|\mathbf{x}) -\gamma(\mathbf{x},y)\right)^2\right]}
\end{equation}
\begin{equation}\label{eq:correctness}\small 
    \alpha(\mathbf{x},y) = \mathbb{E}_t\left[ \mathbb{I}\{y=\underset{y'\in\mathcal{Y}}{\arg\max}\,  p_{\theta_t}(y'|\mathbf{x}) \}\right]
\end{equation}

These three variables map each instance into a three-dimensional continuous space, yielding distinct regions corresponding to diverse sample characteristics. \citeauthor{swayamdipta-etal-2020-dataset}\ named these regions as \emph{easy} (high confidence, low variability), \emph{hard} (low confidence, low variability) and \emph{ambiguous} (high variability) samples, where the latter tend to reside closer to the decision boundary, and would be informative for improving generalization.

To generalize these metrics to different types of NLP tasks, we introduce the concept of the \emph{individual confidence} ($\lambda$) of a sample at timestep $t$. In sequence classification, $\lambda(\mathbf{x},y,t)=p_{\theta_t}(\mathbf{x},y,t)$ 
corresponds to the model’s predicted probability for the true class,
but in other tasks, $\lambda$ is adapted to account for different target structures.
The confidence and variability are computed directly from $\lambda$, by averaging and scaling on the number of training steps $t\in[1,T]$, so $\gamma_*(\cdot)=\mathbb{E}_t[\lambda_*(\cdot, t)]$ and $\sigma_*(\cdot)=\text{std}_t[\lambda_*(\cdot, t)]$.\\

In \textbf{multi-label classification (MLC)}, each training sample is defined as $(\mathbf{x}, \boldsymbol{y})$, where $\boldsymbol{y}\subseteq\mathcal{Y}$ corresponds to a set of target classes. The model $p_{\theta_t}(y_j=1|\mathbf{x})$ estimates the Bernoulli distribution over each $y_j\in\mathcal{Y}$ conditioned on $\mathbf{x}$, so
confidence and variability are reinterpreted by averaging the individual label contributions (Equation \ref{eq:multilabel-individual}), and the correctness is defined with the averaged Jaccard score (Equation \ref{eq:correctness-multilabel}) given a threshold $\tau\in[0,1]$.
\begin{equation}\label{eq:multilabel-individual}\small 
    \lambda_\text{MLC}(\mathbf{x},\boldsymbol{y},t) = \frac{1}{|\boldsymbol{y}|}\sum_{y_j\in\boldsymbol{y}}  
    p_{\theta_t}(y_j=1|\mathbf{x})
\end{equation}

\begin{equation}\label{eq:correctness-multilabel}\small 
\begin{gathered}
    \alpha_\text{MLC}(\mathbf{x,\boldsymbol{y}}) = \frac{1}{T}\sum_{t=1}^T\frac{|\boldsymbol{y}\cap \hat{\boldsymbol{y}}_{\theta_t}|}{|\boldsymbol{y}\cup \hat{\boldsymbol{y}}_{\theta_t}|} \\
    \text{where } \hat{\boldsymbol{y}}_{\theta_t}=\{y_j:     p_{\theta_t}(y_j=1|\mathbf{x}) > \tau \}_{y_j\in\mathcal{Y}}
\end{gathered}
\end{equation}

In \textbf{sequence labeling (SL)}, each training sample is defined as $(\mathbf{x},\mathbf{y})$, where $\mathbf{y}=(y_1\dots y_n)\in\mathcal{Y}^{n}$ associates each token $x_i$ with a label $y_i$. The model $p_{\theta_t}^i$ estimates the categorical distribution over  $\mathcal{Y}$ for the $i$-th label, enabling either averaging (Equation \ref{eq:sl-individual}) or considering individual token contributions (Equation \ref{eq:sl-individual-token}) to compute the confidence, variability, and correctness.
\begin{equation}\label{eq:sl-individual}
    \lambda_\text{SL}(\mathbf{x},\mathbf{y},t) = \frac{1}{n}\sum_{i=1}^n p_{\theta_t}^i(y_i|\mathbf{x})
\end{equation}
\begin{equation}\label{eq:sl-individual-token}
    \lambda^i_\text{SL}(\mathbf{x},y_i,t) =  p_{\theta_t}^i(y_i|\mathbf{x})
\end{equation}

In \textbf{graph prediction (GP)}, each training sample is defined as $(\mathbf{x}, E)$, where $E$ denotes the set of edges forming a labeled directed graph over $\mathbf{x}$.
We represent an edge connecting tokens $x_i$ and $x_j$ through the label $r\in\mathcal{R}$ as $e=(h\overset{r}{\to} i)$. Since modeling $p(E|\mathbf{x})$ is intractable, parameterized models factorize the distribution at edge-level, $p(E|\mathbf{x}):=\prod_e p_{\theta_t}(e|\mathbf{x})$. Their training dynamics can thus be measured by averaging edge-level performance (Equation~\ref{eq:confidence-gp}):
\begin{equation}\label{eq:confidence-gp}\small 
    \lambda_\text{GP}(\mathbf{x},E,t)  = \frac{1}{|E|} \sum_{e\in E} p_{\theta_t}(e|\mathbf{x}) 
\end{equation}
\begin{equation}\label{eq:correctness-gp}\small 
    \begin{gathered}
    \alpha_\text{GP}(\mathbf{x},E)  =\frac{1}{T}\sum_{t=1}^T \frac{|E\cap\hat{E}_{\theta_t}|}{|E\cup \hat{E}_{\theta_t}|} \\
    \text{ where }\hat{E}_{\theta_t}=\{e:p_{\theta_t}(e|\mathbf{x}) > \tau \}
    \end{gathered}
\end{equation}
In the specific case of tree dependency parsing (DP), edge factorization coincides with token factorization since each edge is uniquely associated with a token, allowing the distribution to be parameterized as a SL approach, where the model learns to assign heads to tokens.
This allows us to express
$\lambda_\text{DP}(\mathbf{x},E,t)=\frac{1}{n}\sum_{i=1}^n p_{\theta_t}((h\overset{r}{\to}i)|\mathbf{x})$ as a special
case of $\lambda_\text{SL}$ and reduce the correctness formulation (Equation \ref{eq:correctness-gp}) to the mean of the correct head predictions.

\paragraph{Distributional comparison of training dynamics}
To assess differences across synthetic datasets, models, and relative to human-authored data, we compare their underlying distributions. Operating at the sample level, the maps emerge from the distribution of individual trajectories. By formalizing model dynamics through the variables defined in Equations \ref{eq:confidence}-\ref{eq:correctness} (and variants), a dataset $\mathcal{D}$ can be characterized by the empirical distribution of training dynamics—induced by a model parameterized by $\theta$ trained on $\mathcal{D}$—over a three-dimensional space of confidence, variability, and correctness, denoted as $\mathcal{T} \subset \mathbb{R}^3$. Since the distribution of the training dynamics is unknown and may be highly non-Gaussian and multimodal,\footnote{This may reflect heterogeneous groups of easy, hard, and ambiguous samples rather than a homogeneous distribution.} we avoid imposing parametric assumptions \cite{dror-etal-2018-hitchhikers}. Instead, we quantify distributional differences using the Wasserstein distance \cite{vaserstein-etal-1969-markov}\footnote{Wasserstein distance is used for distribution matching in NLP and vision \cite{laouar-etal-2023-large, arjovsky2017wasserstein}.} as a metric-based measure and Cliff’s delta \cite{cliff-etal-1993-dominance}\footnote{Cliff’s delta takes values in $[-1,1]$ and measures dominance as the probability that a randomly drawn value from one distribution exceeds a value from another; $0$ indicates no systematic difference, while $\pm 1$ indicates complete dominance.} as a non-parametric effect size over confidence, variability, and correctness.

\subsection{Synthetic Data Generation}\label{section-synthetic-data-generation}

Our goal is to characterize the training dynamics across different generative processes and model parameterizations. We distinguish between  (i) $\mathcal{T}_\mathcal{H}^{\tilde{\theta}}$, the training dynamics induced by a human-authored data distribution ($\mathcal{H}$) under a reference model with parameters  $\tilde{\theta}$; and (ii) $\mathcal{T}_{\mathcal{S}}^\theta$, the dynamics of a synthetic distribution ($\mathcal{S}$) produced by an LLM-based process and evaluated under an arbitrary model parameterization $\theta$, where potentially $\theta\neq \tilde{\theta}$. Unless otherwise stated, we assume $\mathcal{H}$ and $\mathcal{S}$ contain the same number of data points.

To assess the utility of different data regimes in the training-dynamics space (\S\ref{sec:expanded-dataset-cartography}), we induce targeted biases into the generative process of $\mathcal{S}$. 
This enables exploration of prompt-based perturbations of $\mathcal{S}$ that steer generation toward regions associated with particular learning signatures in $\mathcal{T}_\mathcal{H}^{\tilde{\theta}}$.
This facilitates a dual analysis: (i) a comparison of synthetic dynamics obtained from different generative processes and model architectures  $\theta$; and (ii) a proxy comparison
of the fidelity of synthetic dynamics relative to the human-authored counterparts.

\paragraph{Generations via prompting} We design four prompting strategies of increasing granularity to generate task-specific samples. (i)  \emph{Naive} provides basic instructions about the target task and the average sequence length of $\mathcal{H}$. (ii) \textit{Exemplified} builds upon the \textit{Naive} baseline by incorporating two few-shot examples uniformly sampled from $\mathcal{H}$.\footnote{The selected examples are fixed for each dataset and reused across all experiments.} (iii) \emph{Difficulty-aware} extends the \textit{Exemplified} approach by including two examples drawn from the ambiguous, easy-, and hard-to-learn partitions of $\mathcal{T}_\mathcal{H}^{\tilde{\theta}}$,\footnote{The reference distribution $\mathcal{T}_\mathcal{H}^{\tilde{\theta}}$, used to inform the \emph{Difficulty-aware} and \emph{Ambiguous-targeted} prompts, was computed with BERT in early experiments.}
without instructing for complexity calibration. (iv) \textit{Ambiguous-targeted} explicitly prompts the model to generate samples resembling those expected in the ambiguous region of  $\mathcal{T}_\mathcal{H}^{\tilde{\theta}}$.\footnote{We include examples categorized by difficulty, as in the \emph{Difficulty-aware} prompt, but explicitly instruct the model to generate ambiguous samples.} By iterating over these strategies, we move from unconstrained synthesis to a taxonomy-informed generation pipeline that isolates specific learning dynamics. Prompt templates are included in Appendix \ref{ap:prompt-design}.

\section{Experimental Framework}

\paragraph{LLM choices}
To generate our synthetic datasets, we rely on open-source LLM families spanning different architectures and scales. We consider Qwen-3 8B \cite{yang2025qwen3technicalreport}, Llama-3.1 8B \cite{llama}, Olmo-3 7B \cite{olmo2026olmo3}, and the Gemma-3 family \cite{gemmateam2025gemma3technicalreport}. The latter is used as a representative case to study the effect of model size, for which we analyze the 4B, 12B, and 27B variants. 
As explained in \S\ref{section-synthetic-data-generation}, we generate multiple versions of the synthetic datasets by conditioning on four prompting strategies: \emph{Naive}, \emph{Exemplified}, \emph{Difficulty-aware}, and \emph{Ambiguous-targeted}. We conducted a preliminary hyperparameter search to tune the top-$k$ and top-$p$ sampling parameters, aiming to balance diversity with fluency and structural consistency across tasks. Appendix~\ref{ap:gen-parameters} reports the selected settings.

\paragraph{Datasets} We consider representative tasks from each NLP paradigm for which training dynamics were defined in \S\ref{sec:expanded-dataset-cartography}.
For single-label classification, we use 
SST-5 \cite{socher-etal-2013-recursive} as a representative sentiment analysis dataset. For multi-label classification, we use the SemEval-2018 Task Emotion Classification (EC) dataset \cite{mohammad-etal-2018-semeval}, which contains Twitter posts annotated with multiple emotion categories.\footnote{We enforce at least one label per synthetic sample to prevent degenerate empty-label annotations}
For SL and GP, we rely on UD 2.15 \cite{nivre-etal-2020-universal}, focusing on universal part-of-speech tagging and dependency parsing on the English EWT treebank.\footnote{For completeness, we further evaluate our framework in multilingual settings (Spanish and French). For SST-5, we use translations generated with NLLB-200 models \cite{nllbteam2022languageleftbehindscaling}; for EC, we use the Spanish split of the dataset (no French split is available); and for SL and GP, we use the Spanish GSD and French Sequoia treebanks. We observe similar behaviors across languages as in English. Full results are reported in Appendix~\ref{ap:multilingual-benchmark}.}

\paragraph{Encoder models for training dynamics}
To compute the empirical training dynamics $\mathcal{T}_{\mathcal{*}}$ associated with a synthetic dataset $\mathcal{S}$ or a human-authored dataset $\mathcal{H}$, we rely on RoBERTa \cite{liu-etal-2020-roberta} as the primary encoder. We  further assess the robustness of the resulting distributions using BERT \cite{devlin-etal-2019-bert}, DistilBERT \cite{sanh-etal-2020-distilbert}, and LLM2Vec\footnote{LLM2Vec converts decoder-only LLMs into encoders by adapting their representations to produce 
embeddings.} \cite{behnamghader-etal-2024-llmvec} with its pre-trained Llama-3.1-8B version as the underlying model and fine-tuning it with LoRA \cite{hu2021loralowrankadaptationlarge}, serving as a proxy for modern autoregressive architectures.

From the encoder representations, we compute task-specific output probabilities. In single-label classification and PoS tagging, we use a linear layer followed by a softmax over the label space, whereas in the multi-label setting, we use element-wise sigmoid activations, producing independent probability estimates for each label. For dependency parsing, we approximate it using scores from the state-of-the-art biaffine parser of \citet{dozat2017deepbiaffineattentionneural} over the product of head–dependent pairs, with separate softmaxes over candidate heads and relations.

\paragraph{Ill-formed LLM outputs} We found that LLM outputs, although well-formed in classification tasks, required some post-processing in structured prediction tasks. In preliminary experiments, we observed that native output formats lead to more formatting issues than JSON-based generation; therefore we adopt the latter to better preserve the intended structure. Also, for PoS tagging, we used Stanza \cite{qi-etal-2020-stanza} to normalize generated labels outside the valid space, correcting an average of $1.64\% \pm 2.89$ of the generated tokens across LLMs and prompts. For dependency parsing, we use simple heuristics to enforce well-formed outputs. In practice, label corrections affect only $1.95\% \pm 1.31$ of dependency relations, whereas most modifications ($28.36\%\pm 9.19$) involve structural fixes to ensure valid trees. This suggests that LLMs struggle to generate fully valid tree structures, which is coherent with recent findings \cite{ezquerro-etal-2025-better}.

\paragraph{Filtering of near-duplicated LLM outputs}
To ensure clean comparisons across distributions, we filter generated samples using a MinHash approximation of Jaccard similarity \cite{broder1997resemblance} with a threshold of $0.4$. Filtering is performed in two stages: (i) reducing redundancy within the synthetic set via a similarity graph and extracting a maximal independent set \cite{luby-1986-simple} to avoid retaining multiple highly similar instances; and (ii) removing samples that are near-duplicates or regurgitations of instances in the human reference dataset used in the prompt.

\paragraph{Evaluation} We report accuracy for SA and PoS tagging, Jaccard score for EC, and labeled attachment score (LAS) for dependency parsing. We use the original test set of each dataset for in-distribution (ID) evaluation and external datasets for out-of-distribution (OOD) evaluation. Specifically, we use Yelp \cite{zhang-etal-2015-yelp} for OOD evaluation on SST-5, and evaluate PoS tagging and dependency parsing on the English PUD test set using models trained on English EWT.\footnote{For SemEval-2018 EC, we do not consider an OOD setting due to the lack of large-scale emotion datasets with a compatible annotation scheme. Existing emotion resources vary significantly in label taxonomies and guidelines, making cross-dataset evaluation confounded by label-space mismatch.}

\section{Results and Analysis}\label{sec:results}

We analyze, both from qualitative and quantitative perspectives, the training dynamics of synthetic datasets (\S \ref{subsec:results-data-quality}); the robustness of the framework for characterizing such samples according to their confidence, variability and correctness (\S\ref{subsec:performance-comparison}); and the impact of exploiting different regions of the resulting data maps during tuning (\S \ref{sec:informed-sample-selection}). Unless otherwise specified, qualitative results correspond to RoBERTa as encoder to produce the data maps, and Gemma-27B as the representative model (being the largest among all considered models) to generate the synthetic datasets with the \emph{Difficulty-aware} prompt strategy (enabling taxonomy-informed synthetic sample generation).

\subsection{Training dynamics distributions}\label{subsec:results-data-quality}

We discuss the data maps derived from synthetic datasets generated by different LLMs across multiple tasks. Since the models were prompted using different strategies, designed to reflect characteristics of reference human datasets (e.g., label space, domain, or task formulation), we analyze how these choices affect the resulting training dynamics and whether they yield patterns similar to those observed in the corresponding human datasets used as references for prompting.

\subsubsection{Human vs. synthetic data maps}\label{ssc:human-cartography-comparison}

\begin{figure}[hptb]\centering
    \includegraphics[width=\linewidth]{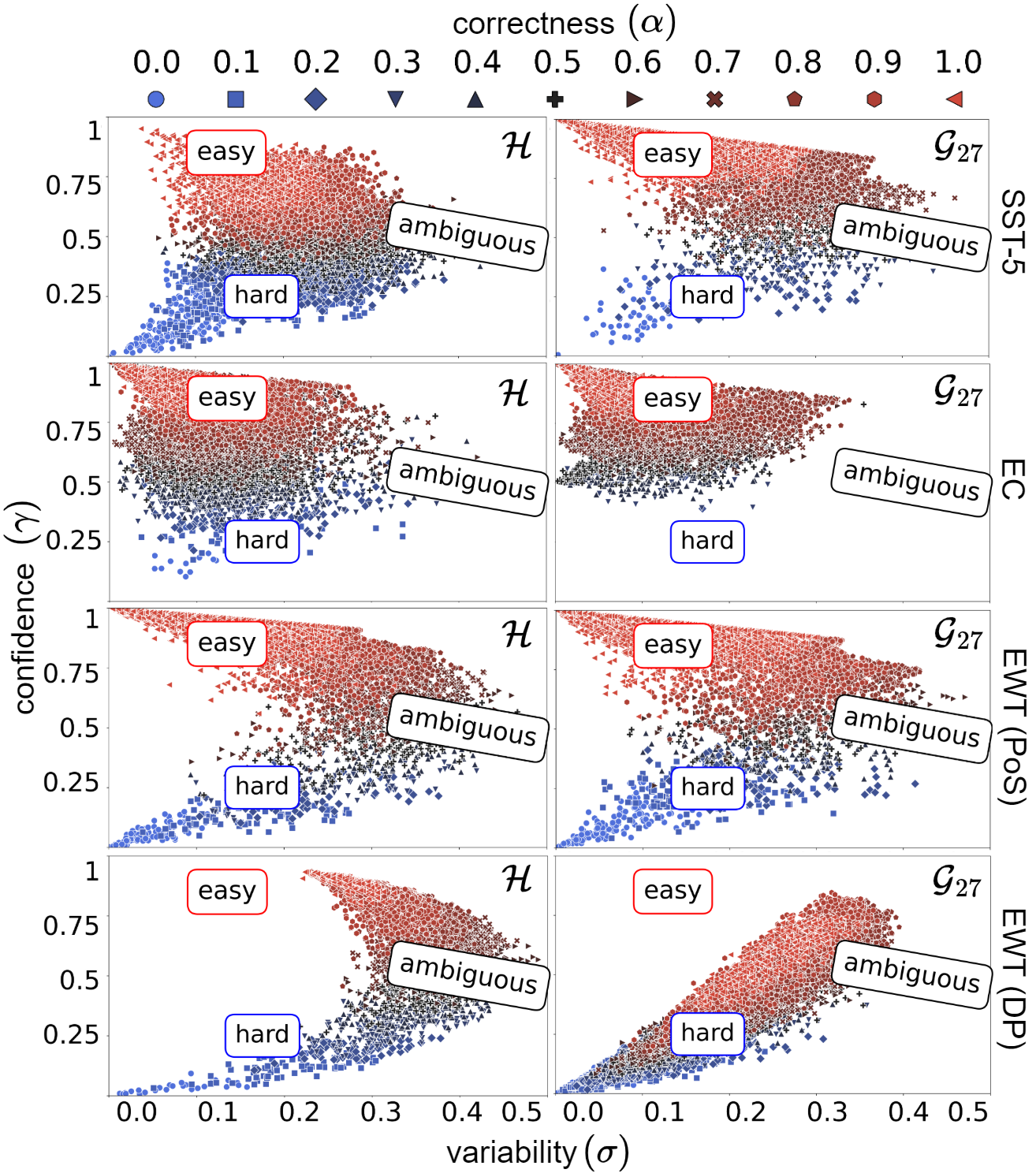}
    \caption{\label{fig:task-cartographies}Human ($\mathcal{H}$) vs. synthetic training dynamics  with Gemma 27B ($\mathcal{G}_\text{27}$) and the \textit{Difficulty-aware} prompt on SST-5 (top), EC (middle-top) and PoS tagging (middle-bottom) and dependency parsing based on EWT (bottom). For display, we subsampled the EWT dataset to 20\,000 uniformly sampled tokens.}
\end{figure}

We first provide a qualitative analysis to ground the subsequent quantitative comparison of training dynamics maps. Figure \ref{fig:task-cartographies} shows the data maps obtained for SST-5 (polarity classification), EC (emotion classification) and EWT (PoS tagging and dependency parsing, at token-level).

The SST-5 and EWT distributions exhibit the characteristic `inverted-C' shape—also observed by \citet{swayamdipta-etal-2020-dataset} for other human-generated datasets—reflecting the relationship between confidence and variability, where high-confidence predictions tend to remain stable across epochs, whereas more uncertain instances display greater variability. EC, instead, shows a concentration of variability in lower ranges, leading to the disappearance of the inverted C-shape.
This effect likely stems from the formulation of multi-label confidence (Equation~\ref{eq:multilabel-individual}). This formulation smooths the model’s predictions across labels, which reduces the sample-level standard deviation.

In qualitative comparison with human counterparts, Gemma 27B reproduces similar training dynamics on sentiment classification and PoS tagging, with comparable easy, hard, and ambiguous regions; but fails to produce hard-to-learn examples for the multi-label EC data. On the other hand, results on dependency parsing highlight difficulties in reproducing human-annotated training dynamics, especially on easy samples, where no sample is characterized by high-confidence and low variability. This suggests a lack of high-quality, learnable synthetic data and is also coherent with previous findings on the limitations of LLMs as zero- or few-shot dependency parsers~\cite{ezquerro-etal-2025-better}.
For parsing, we also verified that the observed training dynamics were not affected by filtering artifacts, as nearly identical distributions are obtained when showing only uncorrected samples (Appendix~\ref{ap:results}).

\subsubsection{Training dynamics across LLMs}\label{ssc:llm-cartography-comparison}

Figure~\ref{fig:llm-comparison-sst5} compares the empirical data maps across the considered LLMs on SST-5. See Appendix~\ref{ap:results} for comparisons on the other datasets.

\begin{figure}[hptb]\centering
    \includegraphics[width=\linewidth]{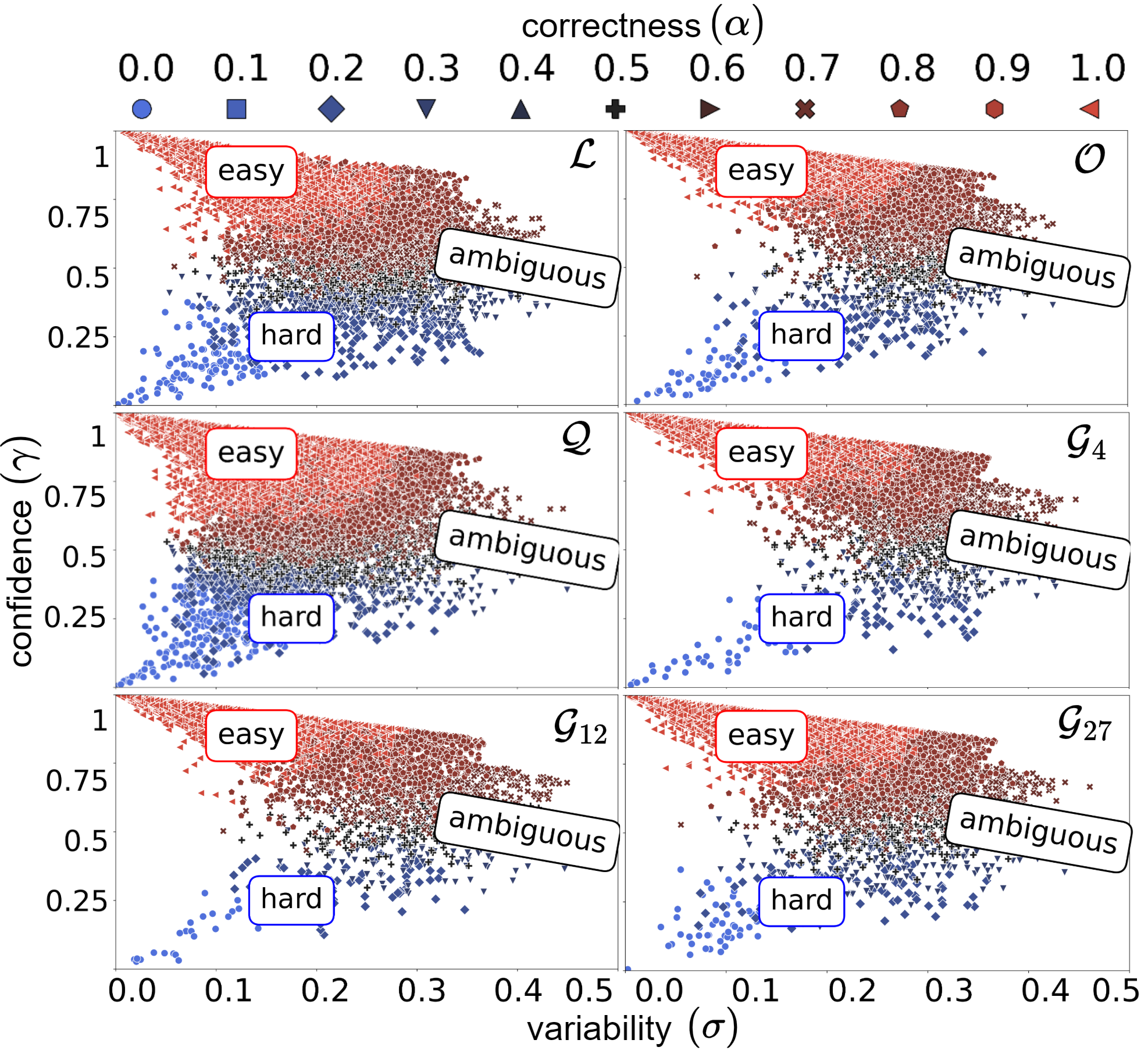}
    \caption{\label{fig:llm-comparison-sst5}Data maps on SA (SST-5) with LLM-generated samples with the \textit{Difficulty-aware} prompt. 
    We include: Gemma 4B (\gemma{4}) Gemma 12B (\gemma{12}) and 27B (\gemma{27}), Llama (\llama), Olmo (\olmo) and Qwen (\qwen).
    }
\end{figure}

\begin{figure}[hptb]\centering
    \includegraphics[width=\linewidth]{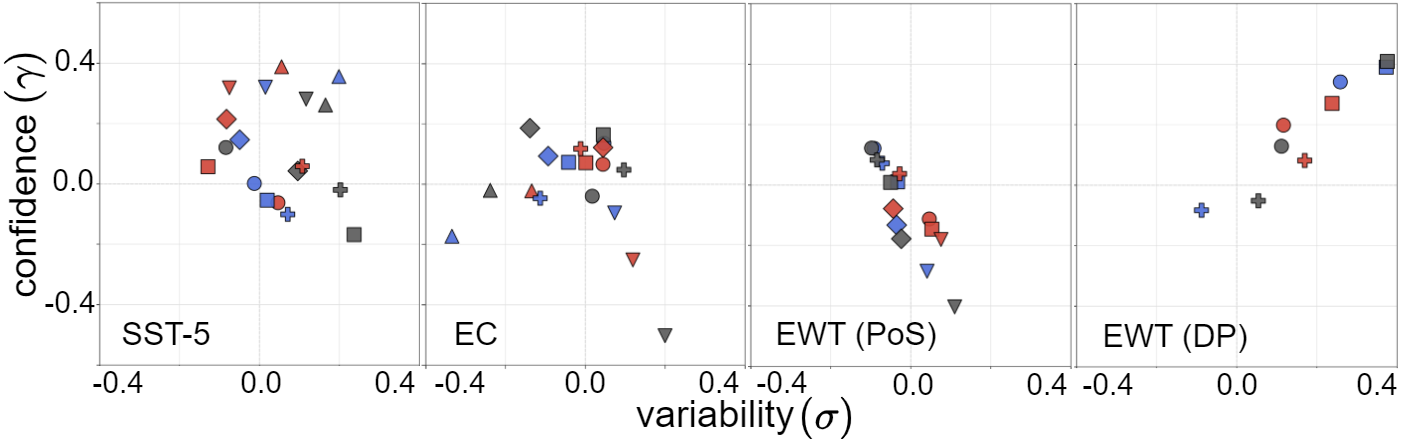}
    \caption{\label{fig:sst5-prompt-comparison}Comparison of training dynamics across prompting strategies. Subfigures illustrate the Cliff's delta for confidence ($\gamma$)  and variability ($\sigma$) across SST-5, EC, EWT (PoS) and EWT (DP) relatively to the \textit{Naive} prompt. Colors denote the prompt strategy (\textcolor[HTML]{d1483a}{$\bullet$} \textit{Exemplified}, \textcolor[HTML]{5070db}{$\bullet$} \textit{Difficulty-aware}, \textcolor[HTML]{5c5c5c}{$\bullet$} \textit{Ambiguous targeted}), while symbols correspond to the LLM used for data generation (\scalebox{0.9}{$\Diamond$}$\mathcal{G}_{4}$, \scalebox{0.7}{\raisebox{2pt}{$\bigcirc$}}$\mathcal{G}_{12}$, \scalebox{0.9}{$\square$}$\mathcal{G}_{27}$, \scalebox{0.8}{$\triangle$}$\mathcal{L}$, \scalebox{1}{$\triangledown$}$\mathcal{O}$).}
\end{figure}

Quantitatively, to summarize the impact of prompting on training dynamics distributions, Figure \ref{fig:sst5-prompt-comparison} presents the distributional shift between the \textit{naive} prompt and the alternative strategies using Cliff's delta on confidence and variability. Our analysis shows some fluctuations in confidence and variability, though without a clear correlation with the prompting complexity. The lack of a consistent pattern (such as increasing divergence with more sophisticated templates) suggests that the training dynamics are insensitive to the specific prompting strategy. Instead, the observed deviations likely stem from the model's generative process.

\begin{figure}[hptb]\centering
    \includegraphics[width=0.95\linewidth]{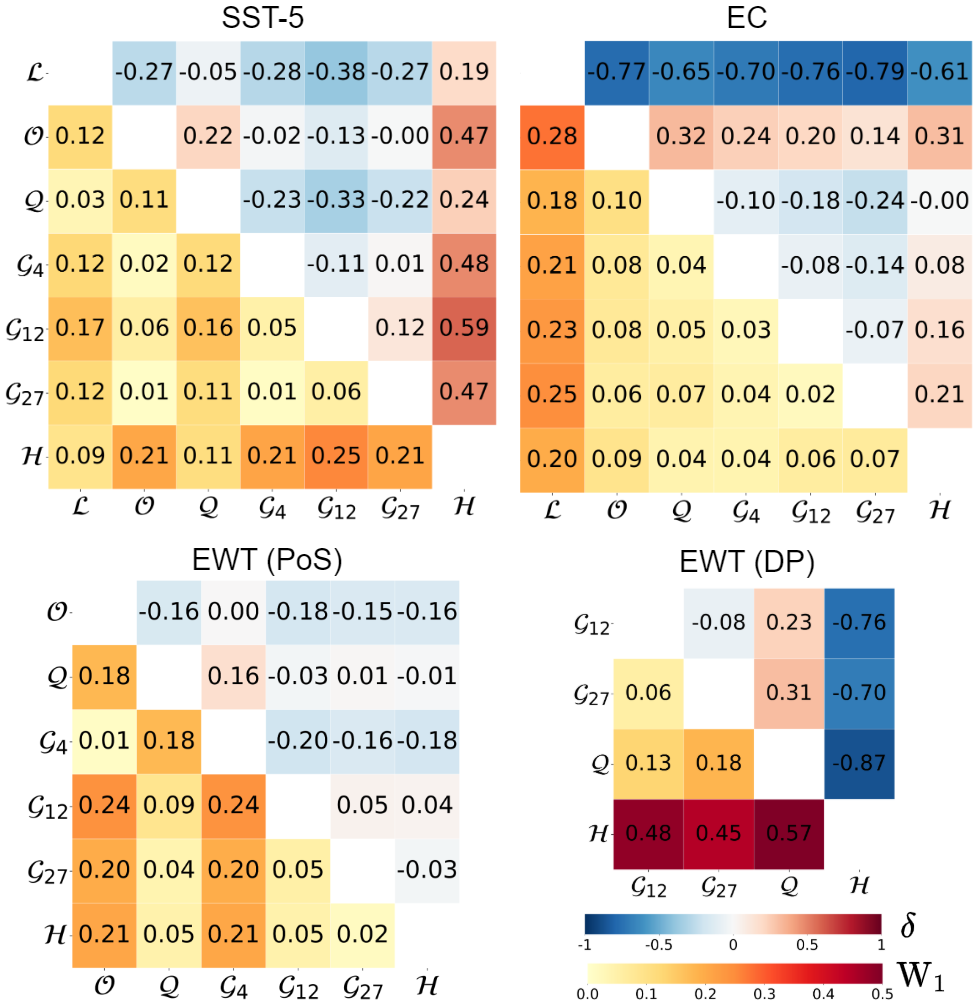}
    \caption{\label{fig:heatmap-comparison}
    Comparison of human and synthetic distributions across LLMs
    and the \textit{Difficulty-aware} prompt using Wasserstein distance on $\mathcal{T}^\theta_*$ (lower triangular matrices), and Cliff's delta, evaluated on correctness (upper triangular matrices). Same notation as in Figure \ref{fig:llm-comparison-sst5}.}
\end{figure}

Figure \ref{fig:heatmap-comparison}\footnote{For PoS tagging and parsing, some models (e.g., LLaMA 3.1 8B or Olmo-3 7B) are omitted because the generation process systematically produced incorrectly formatted outputs, despite prompt-based control efforts.} compares LLM-generated synthetic datasets under the \emph{Difficulty-aware} strategy by quantifying differences in their data maps. We include a human reference row to complete the human vs synthetic comparison (\S\ref{ssc:human-cartography-comparison}). The lower triangles report pairwise Wasserstein distances (\wasserstein), while the upper triangular matrices report Cliff’s delta (\cliffdelta) on correctness.

Overall, \wasserstein distances indicate that training dynamics distributions are more consistent across LLMs than across tasks. In SST-5, LLaMA and Qwen yield highly similar distributions (\wasserstein $= 0.03$), while Olmo closely aligns with the Gemma family (e.g., \wasserstein values between 0.01 and 0.06). In EC, all models except LLaMA produce similar distributions, with consistently small pairwise \wasserstein values. For PoS tagging, the closest agreement is observed between Gemma and Qwen,
while dependency parsing shows substantially larger divergences across models, likely due to inconsistent or poor generation. With respect to the human reference, despite some variation across tasks, Qwen and the Gemma models  generally produce the distributions closest to those of humans, although LLaMA's distribution for SST-5 is also highly aligned.

Cliff’s delta results on correctness further characterize these differences in terms of learnability. As observed in \wasserstein, in SST-5, LLaMA and Qwen exhibit small pairwise effect sizes, indicating similar correctness distributions. EC shows more negative values for LLaMA.
For PoS tagging, Cliff’s delta values are near zero between Gemma and Qwen, suggesting highly similar behavior across synthetic distributions. In contrast, dependency parsing shows much larger discrepancies, with stronger negative effect sizes across model comparisons. Compared to human-authored datasets, most synthetic datasets show positive Cliff’s delta on SST-5 and EC, indicating easier samples, whereas dependency parsing consistently shows strongly negative values, suggesting that synthetic data fails to reproduce even the easy-to-learn regions observed in human training dynamics.

\subsection{Dynamics robustness across encoders}
\label{subsec:performance-comparison}

\begin{figure}[hpbt]\centering
    \includegraphics[width=\linewidth]{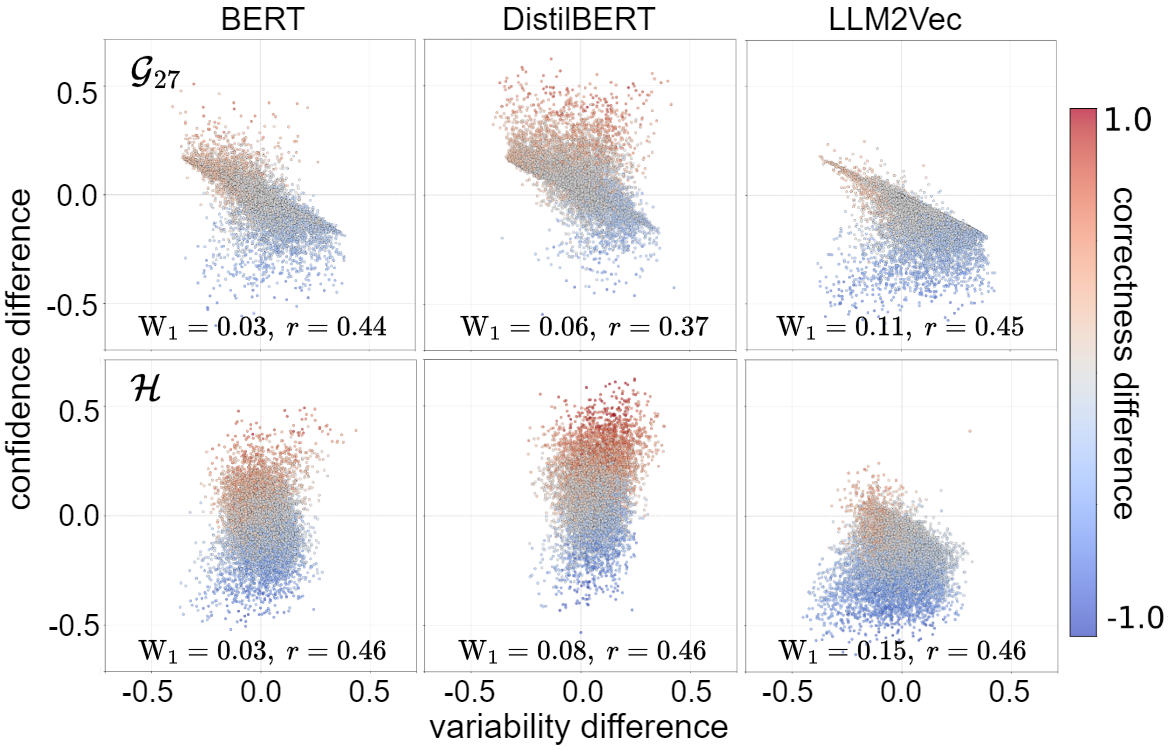}
    \caption{\label{fig:sst5-encoder-comparison}Differences on the training dynamics across encoders with synthetic (top) and human (bottom) data on SST-5. Subfigures show instance-level differences in confidence, variability, and correctness relative to the original RoBERTa encoder used in \S\ref{ssc:human-cartography-comparison}. Here, $r$ denotes the percentage of ambiguous samples (top 33\% highest variability) shared with RoBERTa.}
\end{figure}

We assess the robustness of our framework to encoder choice by analyzing sample-level variations in $\mathcal{T}^\theta_{\mathcal{S}}$ or $\mathcal{T}^\theta_{\mathcal{H}}$ when replacing RoBERTa with DistilBERT, BERT, or LLM2Vec on SST-5 (Figure~\ref{fig:sst5-encoder-comparison}).
The results indicate that the relative positioning of samples within the empirical data maps remains overall stable across encoders: differences in learnability metrics are concentrated around small values. LLM2Vec exhibits a tendency toward negative differences in correctness and confidence, suggesting a better fit on the training set relative to RoBERTa (the reference model). We observe a divergent pattern between synthetic and human data: while human-authored samples seem more evenly distributed across all four quadrants, indicating cross-model consistency; synthetic samples are polarized in the 
second and fourth quadrants, displaying a negative correlation that aligns with a downward-sloping regression fit. This trend suggests that synthetic samples are 
easier or harder for specific architectures to learn, suggesting that synthetic data potentially contains architecture-specific biases that do not generalize as consistently as human annotations. 
Similar analyses for other tasks are included in Appendix \ref{ap:results}.\footnote{This aligns with work on selecting samples for training small models that benefit larger models \cite{du-etal-2025-ftft}, though limited to human data and DeBERTa-scale models.}

\begin{table}[htpb!]
    \scriptsize\centering
    \setlength{\tabcolsep}{1.6pt}
    \renewcommand{\arraystretch}{0.95}
    \begin{tabular}{ll|cccc|ccc}
        \hline
            && \multicolumn{4}{c|}{\textit{in-distribution}} & \multicolumn{3}{c}{\textit{out-of-distribution}}\\
         & & {SST-5} & {EC} & {EWT\textsubscript{PoS}} & {EWT\textsubscript{DP}} & {SST-5} & {EWT\textsubscript{PoS}} & {EWT\textsubscript{DP}} \\
         \hline
         \multicolumn{1}{l}{\parbox[t]{2mm}{\multirow{5}{*}{\rotatebox[origin=c]{90}{\textit{full}}}}} & $\mathcal{L}$ & \cellcolor[RGB]{216,222,163}{40.7} & \cellcolor[RGB]{211,221,163}{38.6} & \nan & \nan & \cellcolor[RGB]{239,225,159}{38.9} & \nan & \nan \\
         & $\mathcal{O}$ & \cellcolor[RGB]{205,219,164}{43.0} & \cellcolor[RGB]{238,212,159}{26.4} & \cellcolor[RGB]{237,202,160}{71.9} & \nan & \cellcolor[RGB]{238,219,159}{37.6} & \cellcolor[RGB]{238,214,159}{74.9} & \nan \\
         & $\mathcal{Q}$ & \cellcolor[RGB]{217,222,162}{40.4} & \cellcolor[RGB]{202,218,165}{40.8} & \cellcolor[RGB]{200,218,165}{82.2} & \cellcolor[RGB]{232,169,167}{8.5} & \cellcolor[RGB]{239,224,159}{38.7} & \cellcolor[RGB]{180,212,167}{86.7} & \cellcolor[RGB]{232,170,167}{8.6} \\
         & $\mathcal{G}_{12}$ & \cellcolor[RGB]{228,225,161}{38.3} & \cellcolor[RGB]{212,221,163}{38.4} & \cellcolor[RGB]{211,221,163}{80.5} & \cellcolor[RGB]{235,179,161}{24.9} & \cellcolor[RGB]{212,221,163}{44.1} & \cellcolor[RGB]{180,212,167}{86.6} & \cellcolor[RGB]{235,183,160}{27.9} \\
         & $\mathcal{H}$ & \cellcolor[RGB]{160,200,159}{58.1} & \cellcolor[RGB]{161,201,159}{59.0} & \cellcolor[RGB]{158,199,158}{93.0} & \cellcolor[RGB]{158,199,158}{93.2} & \cellcolor[RGB]{162,202,160}{57.1} & \cellcolor[RGB]{158,199,158}{94.4} & \cellcolor[RGB]{158,199,158}{92.3} \\
         \hline
         \multicolumn{1}{l}{\parbox[t]{2mm}{\multirow{5}{*}{\rotatebox[origin=c]{90}{\textit{random}}}}} & $\mathcal{L}$ & \cellcolor[RGB]{234,227,160}{37.0} & \cellcolor[RGB]{234,227,160}{32.8} & \nan & \nan & \cellcolor[RGB]{239,224,159}{38.6} & \nan & \nan \\
         & $\mathcal{O}$ & \cellcolor[RGB]{233,174,165}{17.6} & \cellcolor[RGB]{237,204,160}{23.9} & \cellcolor[RGB]{235,181,160}{68.0} & \nan & \cellcolor[RGB]{232,168,168}{20.0} & \cellcolor[RGB]{235,180,160}{68.5} & \nan \\
         & $\mathcal{Q}$ & \cellcolor[RGB]{234,227,160}{37.0} & \cellcolor[RGB]{211,221,163}{38.5} & \cellcolor[RGB]{201,218,165}{82.0} & \cellcolor[RGB]{232,170,167}{8.7} & \cellcolor[RGB]{239,227,159}{39.3} & \cellcolor[RGB]{181,213,168}{86.2} & \cellcolor[RGB]{232,169,167}{7.3} \\
         & $\mathcal{G}_{12}$ & \cellcolor[RGB]{226,224,161}{38.7} & \cellcolor[RGB]{209,220,164}{39.2} & \cellcolor[RGB]{207,220,164}{81.1} & \cellcolor[RGB]{234,178,162}{22.6} & \cellcolor[RGB]{199,217,165}{46.3} & \cellcolor[RGB]{181,212,167}{86.4} & \cellcolor[RGB]{235,180,161}{24.6} \\
         & $\mathcal{H}$ & \cellcolor[RGB]{168,205,162}{54.2} & \cellcolor[RGB]{161,201,159}{58.9} & \cellcolor[RGB]{161,201,159}{92.1} & \cellcolor[RGB]{159,200,159}{92.1} & \cellcolor[RGB]{164,202,160}{56.6} & \cellcolor[RGB]{161,200,159}{93.5} & \cellcolor[RGB]{159,199,158}{91.6} \\
         \hline
         \multicolumn{1}{l}{\parbox[t]{2mm}{\multirow{5}{*}{\rotatebox[origin=c]{90}{\textit{max.$\gamma$}}}}} & $\mathcal{L}$ & \cellcolor[RGB]{233,174,165}{17.6} & \cellcolor[RGB]{215,222,163}{37.7} & \nan & \nan & \cellcolor[RGB]{232,168,168}{20.0} & \nan & \nan \\
         & $\mathcal{O}$ & \cellcolor[RGB]{226,225,161}{38.6} & \cellcolor[RGB]{237,200,160}{22.8} & \cellcolor[RGB]{239,222,159}{75.5} & \nan & \cellcolor[RGB]{236,190,160}{31.6} & \cellcolor[RGB]{219,223,162}{80.4} & \nan \\
         & $\mathcal{Q}$ & \cellcolor[RGB]{237,200,160}{28.9} & \cellcolor[RGB]{188,215,167}{44.4} & \cellcolor[RGB]{195,217,166}{82.8} & \cellcolor[RGB]{232,168,168}{6.5} & \cellcolor[RGB]{233,171,166}{22.1} & \cellcolor[RGB]{181,212,167}{86.4} & \cellcolor[RGB]{232,169,168}{6.2} \\
         & $\mathcal{G}_{12}$ & \cellcolor[RGB]{234,176,163}{19.9} & \cellcolor[RGB]{207,220,164}{39.6} & \cellcolor[RGB]{207,220,164}{81.1} & \cellcolor[RGB]{234,178,162}{22.3} & \cellcolor[RGB]{233,172,166}{23.0} & \cellcolor[RGB]{184,213,168}{85.7} & \cellcolor[RGB]{234,178,162}{22.7} \\
         & $\mathcal{H}$ & \cellcolor[RGB]{164,202,160}{56.5} & \cellcolor[RGB]{173,208,164}{51.5} & \cellcolor[RGB]{164,202,160}{91.1} & \cellcolor[RGB]{164,203,161}{87.5} & \cellcolor[RGB]{162,201,160}{57.3} & \cellcolor[RGB]{159,200,159}{93.9} & \cellcolor[RGB]{162,201,160}{88.7} \\
         \hline
         \multicolumn{1}{l}{\parbox[t]{2mm}{\multirow{5}{*}{\rotatebox[origin=c]{90}{\textit{max.$\alpha$}}}}} & $\mathcal{L}$ & \cellcolor[RGB]{224,224,161}{39.0} & \cellcolor[RGB]{203,219,165}{40.6} & \nan & \nan & \cellcolor[RGB]{227,225,161}{41.5} & \nan & \nan \\
         & $\mathcal{O}$ & \cellcolor[RGB]{228,225,161}{38.3} & \cellcolor[RGB]{237,200,160}{22.8} & \cellcolor[RGB]{239,226,159}{76.2} & \nan & \cellcolor[RGB]{239,222,159}{38.3} & \cellcolor[RGB]{221,223,162}{80.1} & \nan \\
         & $\mathcal{Q}$ & \cellcolor[RGB]{239,226,159}{35.5} & \cellcolor[RGB]{177,210,166}{49.0} & \cellcolor[RGB]{200,218,165}{82.1} & \cellcolor[RGB]{232,169,167}{8.1} & \cellcolor[RGB]{236,191,160}{31.8} & \cellcolor[RGB]{182,213,168}{86.1} & \cellcolor[RGB]{232,169,167}{7.2} \\
         & $\mathcal{G}_{12}$ & \cellcolor[RGB]{208,220,164}{42.4} & \cellcolor[RGB]{203,218,165}{40.7} & \cellcolor[RGB]{208,220,164}{81.0} & \cellcolor[RGB]{234,179,161}{24.1} & \cellcolor[RGB]{191,215,167}{47.6} & \cellcolor[RGB]{180,212,167}{86.5} & \cellcolor[RGB]{235,179,161}{24.0} \\
         & $\mathcal{H}$ & \cellcolor[RGB]{158,199,158}{59.3} & \cellcolor[RGB]{158,199,158}{60.6} & \cellcolor[RGB]{163,202,160}{91.4} & \cellcolor[RGB]{164,203,161}{87.5} & \cellcolor[RGB]{158,199,158}{58.9} & \cellcolor[RGB]{161,200,159}{93.5} & \cellcolor[RGB]{162,202,160}{88.3} \\
         \hline
         \multicolumn{1}{l}{\parbox[t]{2mm}{\multirow{5}{*}{\rotatebox[origin=c]{90}{\textit{max.$\sigma$}}}}} & $\mathcal{L}$ & \cellcolor[RGB]{233,174,165}{17.6} & \cellcolor[RGB]{218,222,162}{36.8} & \nan & \nan & \cellcolor[RGB]{232,168,168}{20.0} & \nan & \nan \\
         & $\mathcal{O}$ & \cellcolor[RGB]{228,225,161}{38.2} & \cellcolor[RGB]{238,214,159}{27.2} & \cellcolor[RGB]{236,198,160}{71.2} & \nan & \cellcolor[RGB]{237,200,160}{33.7} & \cellcolor[RGB]{237,204,160}{73.2} & \nan \\
         & $\mathcal{Q}$ & \cellcolor[RGB]{235,227,160}{36.8} & \cellcolor[RGB]{220,223,162}{36.2} & \cellcolor[RGB]{202,218,165}{81.9} & \cellcolor[RGB]{232,169,167}{8.5} & \cellcolor[RGB]{238,219,159}{37.6} & \cellcolor[RGB]{181,213,168}{86.2} & \cellcolor[RGB]{232,169,167}{7.4} \\
         & $\mathcal{G}_{12}$ & \cellcolor[RGB]{225,224,161}{38.9} & \cellcolor[RGB]{209,220,164}{39.0} & \cellcolor[RGB]{210,220,164}{80.7} & \cellcolor[RGB]{234,178,162}{22.4} & \cellcolor[RGB]{208,220,164}{44.7} & \cellcolor[RGB]{182,213,168}{85.9} & \cellcolor[RGB]{235,179,161}{23.8} \\
         & $\mathcal{H}$ & \cellcolor[RGB]{174,208,165}{51.6} & \cellcolor[RGB]{165,203,161}{56.6} & \cellcolor[RGB]{161,201,159}{92.1} & \cellcolor[RGB]{159,199,158}{92.6} & \cellcolor[RGB]{173,207,164}{53.0} & \cellcolor[RGB]{162,201,160}{93.1} & \cellcolor[RGB]{158,199,158}{92.1} \\
         \hline
         \multicolumn{1}{l}{\parbox[t]{2mm}{\multirow{5}{*}{\rotatebox[origin=c]{90}{\textit{mid.$\alpha$}}}}} & $\mathcal{L}$ & \cellcolor[RGB]{239,228,159}{35.9} & \cellcolor[RGB]{213,221,163}{38.2} & \nan & \nan & \cellcolor[RGB]{202,218,165}{45.8} & \nan & \nan \\
         & $\mathcal{O}$ & \cellcolor[RGB]{224,224,161}{39.0} & \cellcolor[RGB]{238,222,159}{29.6} & \cellcolor[RGB]{233,171,166}{61.7} & \nan & \cellcolor[RGB]{211,221,163}{44.2} & \cellcolor[RGB]{232,170,167}{61.9} & \nan \\
         & $\mathcal{Q}$ & \cellcolor[RGB]{231,226,160}{37.6} & \cellcolor[RGB]{230,226,160}{33.8} & \cellcolor[RGB]{206,219,164}{81.3} & \cellcolor[RGB]{232,169,167}{7.6} & \cellcolor[RGB]{237,205,159}{34.7} & \cellcolor[RGB]{184,213,168}{85.7} & \cellcolor[RGB]{232,169,167}{6.9} \\
         & $\mathcal{G}_{12}$ & \cellcolor[RGB]{238,214,159}{32.4} & \cellcolor[RGB]{231,226,160}{33.6} & \cellcolor[RGB]{202,218,165}{81.8} & \cellcolor[RGB]{235,179,161}{24.9} & \cellcolor[RGB]{215,222,163}{43.5} & \cellcolor[RGB]{180,212,167}{86.8} & \cellcolor[RGB]{235,180,160}{25.6} \\
         & $\mathcal{H}$ & \cellcolor[RGB]{178,210,166}{49.8} & \cellcolor[RGB]{174,208,165}{50.8} & \cellcolor[RGB]{164,202,160}{91.1} & \cellcolor[RGB]{159,199,158}{92.5} & \cellcolor[RGB]{179,211,167}{50.3} & \cellcolor[RGB]{165,203,161}{92.1} & \cellcolor[RGB]{158,199,158}{92.0} \\
         \hline
         \multicolumn{1}{l}{\parbox[t]{2mm}{\multirow{5}{*}{\rotatebox[origin=c]{90}{\textit{min.$\gamma$}}}}} & $\mathcal{L}$ & \cellcolor[RGB]{229,225,161}{37.9} & \cellcolor[RGB]{232,168,168}{2.3} & \nan & \nan & \cellcolor[RGB]{237,210,159}{35.7} & \nan & \nan \\
         & $\mathcal{O}$ & \cellcolor[RGB]{232,168,168}{12.6} & \cellcolor[RGB]{238,221,159}{29.2} & \cellcolor[RGB]{235,180,161}{67.4} & \nan & \cellcolor[RGB]{232,168,168}{20.0} & \cellcolor[RGB]{235,181,160}{68.7} & \nan \\
         & $\mathcal{Q}$ & \cellcolor[RGB]{233,174,164}{18.1} & \cellcolor[RGB]{232,226,160}{33.2} & \cellcolor[RGB]{202,218,165}{81.8} & \cellcolor[RGB]{233,171,166}{11.6} & \cellcolor[RGB]{232,168,168}{20.0} & \cellcolor[RGB]{182,213,168}{85.9} & \cellcolor[RGB]{233,171,166}{9.8} \\
         & $\mathcal{G}_{12}$ & \cellcolor[RGB]{231,226,160}{37.6} & \cellcolor[RGB]{217,222,163}{37.2} & \cellcolor[RGB]{207,219,164}{81.2} & \cellcolor[RGB]{234,178,162}{22.4} & \cellcolor[RGB]{180,212,167}{49.8} & \cellcolor[RGB]{179,211,167}{86.9} & \cellcolor[RGB]{235,179,161}{24.1} \\
         & $\mathcal{H}$ & \cellcolor[RGB]{238,217,159}{33.2} & \cellcolor[RGB]{172,207,164}{52.4} & \cellcolor[RGB]{164,202,160}{91.0} & \cellcolor[RGB]{159,199,158}{92.6} & \cellcolor[RGB]{237,204,160}{34.4} & \cellcolor[RGB]{164,203,161}{92.2} & \cellcolor[RGB]{158,199,158}{92.1} \\
         \hline
         \multicolumn{1}{l}{\parbox[t]{2mm}{\multirow{5}{*}{\rotatebox[origin=c]{90}{\textit{min.$\alpha$}}}}} & $\mathcal{L}$ & \cellcolor[RGB]{235,180,161}{23.1} & \cellcolor[RGB]{236,191,160}{20.1} & \nan & \nan & \cellcolor[RGB]{232,168,168}{20.0} & \nan & \nan \\
         & $\mathcal{O}$ & \cellcolor[RGB]{229,225,161}{37.9} & \cellcolor[RGB]{238,219,159}{28.7} & \cellcolor[RGB]{232,168,168}{60.0} & \nan & \cellcolor[RGB]{235,180,161}{28.9} & \cellcolor[RGB]{232,168,168}{60.5} & \nan \\
         & $\mathcal{Q}$ & \cellcolor[RGB]{238,217,159}{33.3} & \cellcolor[RGB]{239,226,159}{30.7} & \cellcolor[RGB]{207,220,164}{81.1} & \cellcolor[RGB]{232,168,168}{6.0} & \cellcolor[RGB]{238,218,159}{37.3} & \cellcolor[RGB]{186,214,167}{85.4} & \cellcolor[RGB]{232,168,168}{5.1} \\
         & $\mathcal{G}_{12}$ & \cellcolor[RGB]{209,220,164}{42.0} & \cellcolor[RGB]{231,226,160}{33.5} & \cellcolor[RGB]{206,219,164}{81.3} & \cellcolor[RGB]{234,174,164}{16.9} & \cellcolor[RGB]{189,215,167}{48.0} & \cellcolor[RGB]{181,212,168}{86.3} & \cellcolor[RGB]{234,175,163}{17.4} \\
         & $\mathcal{H}$ & \cellcolor[RGB]{237,205,159}{30.2} & \cellcolor[RGB]{181,213,168}{46.5} & \cellcolor[RGB]{163,202,160}{91.3} & \cellcolor[RGB]{160,200,159}{91.1} & \cellcolor[RGB]{238,214,159}{36.5} & \cellcolor[RGB]{164,203,161}{92.2} & \cellcolor[RGB]{159,200,159}{91.1} \\
         \hline
    \end{tabular}
    \caption{\label{fig:subsets_roberta}Performance on full data and sample subsets of LLMs selected via uniform sampling: uniform selection (\textit{random}), maximizing (\textit{max}) or minimizing (\textit{min}) confidence ($\gamma$), correctness ($\alpha$) and variability ($\sigma$); and approximating the variability threshold (\textit{mid.}$\alpha$). The last row ($\mathcal{H}$) corresponds to the performance with the human-counterpart dataset. Colors are independently normalized per column on a continuous scale from low (red) to high (green) performance to visually illustrate trends. RoBERTa-large is used both to compute $\alpha$, $\gamma$, and $\sigma$; and to train on the selected subsets.}
\end{table}

\subsection{Training with selected data}\label{sec:informed-sample-selection}

Finally, we study how training on different data strata affects performance. We limit each subset to 33\% of the dataset maximizing confidence, variability, or correctness metrics \cite{swayamdipta-etal-2020-dataset};
and consider two baselines: 
(i) a 33\% random subset baseline, and (ii) a full-dataset model.
Table \ref{fig:subsets_roberta} shows that the behavior varies across tasks and subsets. Synthetic subsets with high correctness (\textit{max}.$\alpha$) tend to achieve stronger and more consistent performance, especially under OOD evaluation. In contrast, subsets dominated by difficult samples (\textit{min}.$\gamma$, \textit{min}.$\alpha$) and by easy samples with high confidence (\textit{max}.$\gamma$) generally obtain weaker results.
Clear differences are observed between synthetic and human data, where models built on top of human data achieve strong performance even with easy samples with \textit{max}.$\gamma$ in both ID and OOD evaluation, indicating that these subsets are still informative for training, as noted in \S\ref{ssc:human-cartography-comparison}.

\section{Conclusion}

We introduce a framework for characterizing LLM-generated data using training dynamics and learnability metrics from encoder-only models. We show that LLM–dataset pairs across families and scales exhibit systematic differences in confidence and variability, while prompting has no consistent effect on these properties. The framework is stable across encoders but reveals distinct patterns between human and synthetic data. Finally, we find that data selection based on these signals impacts performance differently than for human data.

\section*{Limitations}

\paragraph{Challenges in data generation} Even for sentiment analysis tasks, where synthetic data generation might appear straightforward, we encountered several practical challenges. First, some LLMs struggled to consistently follow the required output format when generating large numbers of samples. Second, we observed a high degree of near-duplication, both among generated samples and with respect to reference human datasets. We mitigate these issues using filtering mechanisms based on Jaccard similarity and BERT-based similarity measures. To match the size of the human dataset, we often had to generate 3–25× more data due to filtering and reduced output diversity under lower generation parameter settings in PoS tagging and dependency parsing.

\paragraph{Synthetic vs.\ human annotation equivalence} Synthetic annotations may not fully capture all nuances of human labeling, particularly in subjective tasks. Nonetheless, although this is difficult to verify reliably, we rely on well-defined benchmarks with relatively high annotation consistency, which reduces ambiguity in label interpretation. Our goal is not to fully replicate human annotation behavior, but rather to study comparative patterns under controlled and scalable annotation regimes.

\paragraph{Robustness of prompts and model choice} We evaluate the stability of our findings across multiple prompt variants and encoder backbones. Empirically, we observe consistent trends across different prompt formulations and models, suggesting that the reported effects are not driven by a specific prompting strategy or encoder choice. This indicates that our results are robust to moderate variations in both generation and representation components. While this does not exclude residual sensitivity to prompt or model design, it provides evidence that the observed patterns reflect general properties of the studied training dynamics rather than idiosyncrasies of a particular configuration.

\paragraph{Limited computational resources} The infrastructure available to us is constrained along several dimensions. We have access to a group server with three RTX A6000 Ada Generation GPUs, one RTX 5000 GPU, and eight RTX 3090 GPUs. In addition, we can access a shared supercomputing facility through a queuing system, providing access to approximately 400 GPUs. Also, due to the computational cost and non-negligible API expenses associated with proprietary models, we focus on open-weight alternatives.

\section*{Ethical Considerations}

Model-generated annotations may reflect or amplify biases present in the underlying models and may not fully align with human annotation standards. Although this is, to some extent, aligned with the motivation of our work, we recommend treating synthetic labels as approximations rather than ground truth.

\paragraph{Use of AI assistants} We used AI assistants to improve the writing quality of the paper in terms of grammar, vocabulary, and phrasing.

\section*{Acknowledgements}
This work was funded  by GAP (PID2022-139308OA-I00) funded by MICIU/AEI/10.13039/501100011033/ and by ERDF, EU; LATCHING (PID2023-147129OB-C21) funded by MICIU/AEI/10.13039/501100011033 and ERDF;  by Xunta de Galicia (ED431C 2024/02); by TSI-100925-2023-1 funded by Ministry for Digital Transformation and Civil Service and “NextGenerationEU” PRTR; and Centro de Investigación de Galicia ‘‘CITIC’’, funded by the Xunta de Galicia through the collaboration agreement between the Consellería de Cultura, Educación, Formación Profesional e Universidades and the Galician universities for the reinforcement of the research centres of the Galician University System (CIGUSA). 

This research project was made possible through the access granted by the Galician Supercomputing Center (CESGA) to its supercomputing infrastructure. The supercomputer FinisTerrae III and its permanent data storage system have been funded by the NextGeneration EU 2021 Recovery, Transformation and Resilience Plan, ICT2021-006904, and also from the Pluriregional Operational Programme of Spain 2014-2020 of the European Regional Development Fund (ERDF), ICTS-2019-02-CESGA-3, and from the State Programme for the Promotion of Scientific and Technical Research of Excellence of the State Plan for Scientific and Technical Research and Innovation 2013-2016 State subprogramme for scientific and technical infrastructures and equipment of ERDF, CESG15-DE-3114. 

We also acknowledge the contribution of the Austrian Science Fund (FWF) [10.55776/COE12].

\bibliography{custom}

\appendix

\section{Synthetic Data Generation}\label{ap:data-generation}

In this appendix, we provide details of our procedure for generating synthetic samples for the selected tasks. We detail the task-specific prompting formats (\S\ref{ap:prompt-design}), strategies (\S\ref{ap:prompt-strategies}), and hyperparameters (\S\ref{ap:gen-parameters}).

\subsection{Prompt design}\label{ap:prompt-design}
In this work we considered four types of tasks: single-label classification, multi-label classification, token-level classification, and tree prediction. We designed the prompt template to satisfy their requirements and adjust to the format of the original datasets. For SST-5 (single-label classification), 
we designed the prompt template to parse outputs in a tabular, comma-separated format. To ensure a balanced target distribution, samples were generated conditioned on a target label. For EC (multi-label emotion classification), we followed a similar approach, generating samples conditioned on a unique label, optionally expanded for diverse label combinations. For PoS tagging and dependency parsing, we use a JSON format with fields adapted to CoNLL-U terminology. Figure \ref{synthetic-samples} shows synthetic samples obtained for each task.

\subsection{Prompt strategies}\label{ap:prompt-strategies}
To analyze whether the prompt can shape data generation differently in terms of training dynamics, we designed four prompting strategies of increasing granularity.

\paragraph{\textit{Naive} prompt} As a first strategy, we only provide basic instructions regarding the task and the average sequence length estimated from the original dataset (Figure \ref{fig:naive-prompt}).

\paragraph{\textit{Exemplified} prompt} The \textit{Naive} strategy is extended by incorporating two reference examples from the original dataset (Figure \ref{fig:exemplified-prompt}), selected based on their proximity to the average sequence length. For PoS tagging and parsing, these reference examples are selected by maximizing label diversity. The same examples are used across all LLMs to ensure consistency.

\paragraph{\textit{Difficulty-aware} prompt} Following the same structure as the \textit{Exemplified} prompt, reference examples are selected from the subsets defined by the training dynamics (easy, ambiguous, and hard) (Figure \ref{fig:difficulty-aware-prompt}).

\paragraph{\textit{Ambiguous-targeted} prompt} This strategy also relies on cartography-based examples, but explicitly instructs the LLM to generate ambiguous samples (Figure \ref{fig:ambiguous-prompt}).

\subsection{Generation parameters}\label{ap:gen-parameters}

Generation parameters were adjusted depending on the LLM and the task. In particular, for PoS tagging and dependency parsing tasks, lower sampling values were required to generate valid UD labels and maintain the correct output format. The selected values are reported in Table \ref{tab:parameters}.

\begin{table}[h!]
    \scriptsize\centering
    \setlength{\tabcolsep}{3pt}
    \begin{tabular}{l|ccc|ccc}
        \hline
        & \multicolumn{3}{c|}{SA} 
        & \multicolumn{3}{c}{PoS/DP} \\
        \cline{2-7}
        & temperature & top-$p$ & top-$k$ & temperature & top-$p$ & top-$k$ \\
        \hline
        \llama                    & 0.8 & 0.95 & 50 & \nan & \nan & \nan \\
        \olmo                     & 0.8 & 0.95 & 50 & 0.5 & 0.9 & 20 \\
        \qwen                     & 1.2 & 0.95 & 50 & 1 & 0.9 & 20 \\
        \gemma{4}  & 1 & 0.95 & 50 & 0.7 & 0.9 & 20 \\
        \gemma{12} & 1 & 0.95 & 50 & 0.8 & 0.9 & 20 \\
        \gemma{27} & 1 & 0.95 & 50 & 0.8 & 0.9 & 20 \\
        \hline
    \end{tabular}
    \caption{LLM configuration for prompting.}
    \label{tab:parameters}
\end{table}

\begin{figure*}[h!]\centering
    \begin{subfigure}[t]{\linewidth}
        \caption{Samples generated with Gemma 3 27B for the SST-5 dataset.}
        \begin{tcolorbox}\begin{lstlisting}
"It wasn't terrible, just... forgettable. A perfectly serviceable piece of entertainment", 2
"Left me feeling optimistic about humanity -  rare quality these days. Highly recommend experiencing it yourself!", 4
        \end{lstlisting}\end{tcolorbox}
    \end{subfigure}
    \begin{subfigure}[t]{\linewidth}
        \caption{Samples generated with Gemma 3 27B for the EC dataset.}
    \begin{tcolorbox}\begin{lstlisting}
"I can't believe the audacity of that company to send me a broken product. What a waste of money.", [0,2]
"Took a pottery class and actually made something recognizable! It's wonky, but I'm proud.", [4,6]
    \end{lstlisting}\end{tcolorbox}
    \end{subfigure}
    \begin{subfigure}[t]{\linewidth}
        \caption{Samples generated with Gemma 3 27B for the EWT dataset (PoS tagging).}
        \begin{tcolorbox}\begin{lstlisting}      
{
  "sentences": [
    {
      "text": "Honestly , I think the AI generated art is getting creepily good, it's almost unsettling.",
      "tokens": ["Honestly", ",", "I", "think", "the", "AI", "generated", "art", "is", "getting", "creepily", "good", ",", "it", "'s", "almost", "unsettling", "."],
      "tags": ["INTJ", "PUNCT", "PRON", "VERB", "DET", "NOUN", "VERB", "NOUN", "AUX", "VERB", "ADV", "ADJ", "PUNCT", "PRON", "AUX", "ADV", "ADJ", "PUNCT"]
    },
    {
      "text": "The new phone's camera is surprisingly good, especially in low light conditions.",
      "tokens": ["The", "new", "phone", "'s", "camera", "is", "surprisingly", "good", ",", "especially", "in", "low", "light", "conditions", "."],
      "tags": ["DET", "ADJ", "NOUN", "PART", "NOUN", "AUX", "ADV", "ADJ", "PUNCT", "ADV", "ADP", "ADJ", "ADJ", "NOUN", "PUNCT"]
    }
  ]
}   \end{lstlisting}\end{tcolorbox}
    \end{subfigure}
    \begin{subfigure}[t]{\linewidth}
        \caption{Samples generated with Gemma 3 27B for the EWT dataset (dependency parsing).}
        \begin{tcolorbox}\begin{lstlisting} 
{
  "sentences": [
    {
      "text": "The children happily built a magnificent sandcastle on the beach .",
      "tokens": ["The", "children", "happily", "built", "a", "magnificent", "sandcastle", "on", "the", "beach", "."],
      "heads": [2, 4, 4, 0, 7, 7, 4, 10, 10, 4, 4],
      "relations": ["det", "nsubj", "advmod", "root", "det", "amod", "obj", "case", "det", "obl", "punct"]
    },
    {
      "text": "The cafe owner quickly resolved the dispute between the two disgruntled customers .",
      "tokens": ["The", "cafe", "owner", "quickly", "resolved", "the", "dispute", "between", "the", "two", "disgruntled", "customers", "."],
      "heads": [3, 3, 5, 5, 0, 7, 5, 12, 12, 12, 12, 5, 5],
      "relations": ["det", "compound", "nsubj", "advmod", "root", "det", "obj", "case", "det", "nummod", "amod", "obl", "punct"]
    }
  ]
}    \end{lstlisting}\end{tcolorbox}
    \end{subfigure}
    \caption{\label{synthetic-samples}Examples from our synthetic datasets.}
\end{figure*}

\begin{figure*}[h!]
    \begin{subfigure}[t]{\linewidth}
        \caption{\label{fig:naive-prompt}\textit{Naive} prompt for the SST-5 dataset.}
        \begin{tcolorbox}\begin{lstlisting}
You are an assistant that generates a synthetic dataset in English composed by film reviews. You write an only csv inside a code block starting with ```csv and ending with ```, without any headers or titles. Don't write any additional text outside the csv. The label is a number, from 0 to the number of classes. 

The review is a film review. In the csv that you have to write, on each different line, it goes first the review (must be in double quotes) and then the label (must be an integer). Don't include more than one review per line. 

The classes in the dataset are very negative (0), negative (1), neutral (2), positive (3) and very positive (4), but now you have to write only 20 film reviews of the class very positive (4), in csv format (review, label). The label must be the integer 4. 

The original dataset has a median length of 19.14 and a standard deviation of 9.31. Try to maintain approximately the same length in your generated sequences.

Generate exclusively novel and distinct samples. Strictly avoid generic phrasing, or repetitive patterns.
    \end{lstlisting}\end{tcolorbox}
    \end{subfigure}
    
    \begin{subfigure}[t]{\linewidth}
        \caption{\label{fig:exemplified-prompt}\textit{Exemplified} prompt for the SST-5 dataset.}
        \begin{tcolorbox}
        \begin{lstlisting}
[naive-prompt]
2 examples of the class very positive (4):
"a stirring, funny and finally transporting re-imagining of beauty and the beast and 1930s horror films", 4
"the film is moody, oozing, chilling and heart-warming all at once ... a twisting, unpredictable, cat-and-mouse thriller.", 4
    \end{lstlisting}\end{tcolorbox}
    \end{subfigure}
    \begin{subfigure}[t]{\linewidth}
        \caption{\label{fig:difficulty-aware-prompt}\textit{Difficulty-aware} prompt for the SST-5 dataset.}
        \begin{tcolorbox}\begin{lstlisting}
[naive-prompt]

2 EASY examples (model consistently predicts correctly):
"a beautifully tooled action thriller about love and terrorism in korea.", 4
"this beautifully animated epic is never dull.", 4

2 AMBIGUOUS examples (model sometimes predicts correctly):
"returning director rob minkoff ... and screenwriter bruce joel rubin ... have done a fine job of updating white's dry wit to a new age.", 4
"a powerful, chilling, and affecting study of one man's dying fall.", 4

2 HARD examples (model consistently predicts incorrectly):
"holofcener rejects patent solutions to dramatize life's messiness from inside out, in all its strange quirks.", 4
"it is sentimental but feels free to offend, is analytical and then surrenders to the illogic of its characters, is about grief and yet permits laughter.", 4
        \end{lstlisting}\end{tcolorbox}
    \end{subfigure}
    \begin{subfigure}[t]{\linewidth}
    \caption{\label{fig:ambiguous-prompt}\textit{Ambiguous-targeted} prompt for the SST-5 dataset.}
        \begin{tcolorbox}\begin{lstlisting}
[difficulty-aware-prompt]
Generate only AMBIGUOUS sequences that are challenging for the model but not impossible to classify correctly.
        \end{lstlisting}
        \end{tcolorbox}
    \end{subfigure}
    \caption{\label{fig:prompt-examples}Prompt templates. Note that the \textit{exemplified} (Figure \ref{fig:exemplified-prompt}) and \textit{Difficulty-aware} (Figure \ref{fig:difficulty-aware-prompt}) extend the \textit{naive} prompt (Figure \ref{fig:naive-prompt}); and the \textit{ambiguous-targeted} (Figure \ref{fig:ambiguous-prompt}) extends the \textit{Difficulty-aware} prompt.}
\end{figure*}

\section{Results on English datasets}\label{ap:results}
We report the complete results of our framework in terms of dataset cartographies and performance on evaluation sets. Due to space limitations, the main content of the paper focused on SST-5 (Section \ref{sec:results}).

\begin{figure}[h!]\centering
    \includegraphics[width=\linewidth]{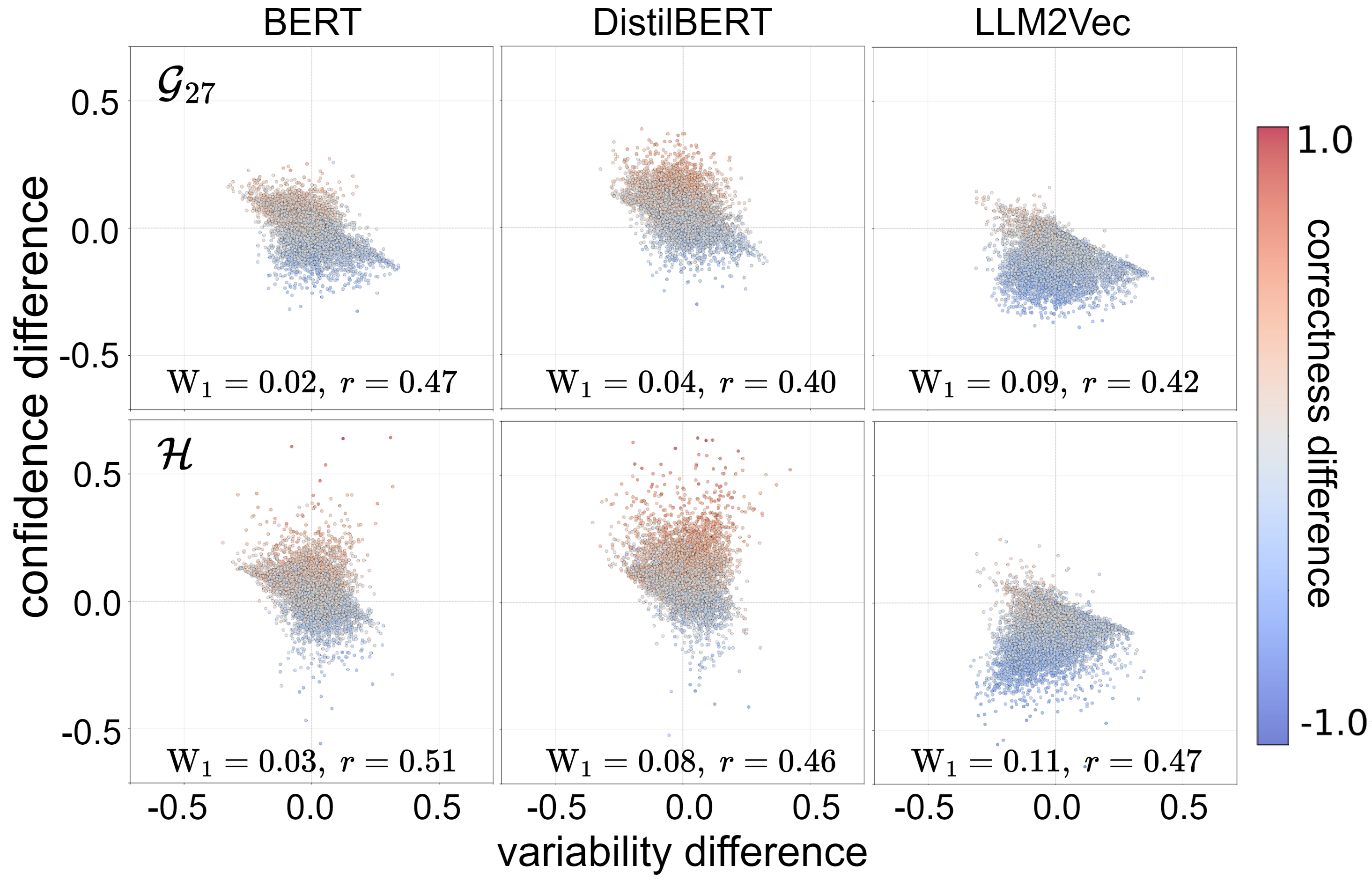}
    \caption{\label{fig:ec-encoder-comparison}Differences on the training dynamics across encoders with synthetic (top) and human (bottom) data on EC. Same notation as in Figure \ref{fig:sst5-encoder-comparison}.}
\end{figure}

\begin{figure}[h!]\centering
    \includegraphics[width=\linewidth]{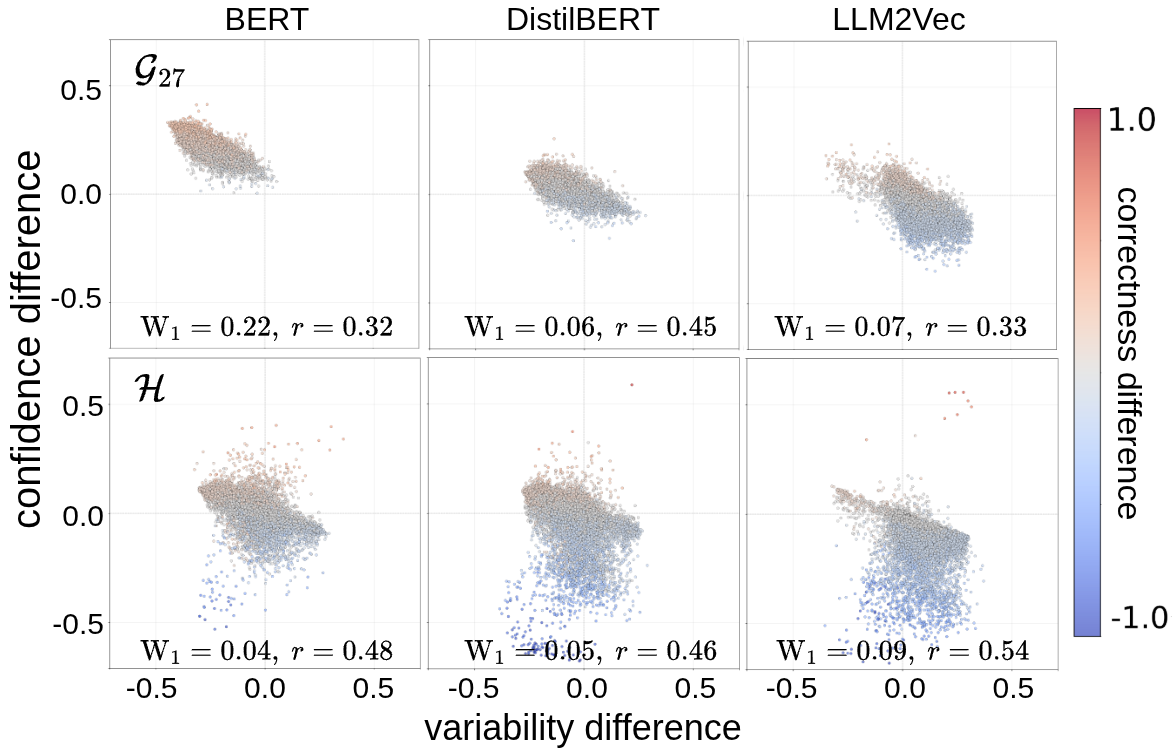}
    \caption{\label{fig:pos-encoder-comparison}Differences on the training dynamics across encoders with synthetic (top) and human (bottom) data on PoS tagging. Same notation as in Figure \ref{fig:sst5-encoder-comparison}.}
\end{figure}

\begin{figure}[h!]\centering
    \includegraphics[width=\linewidth]{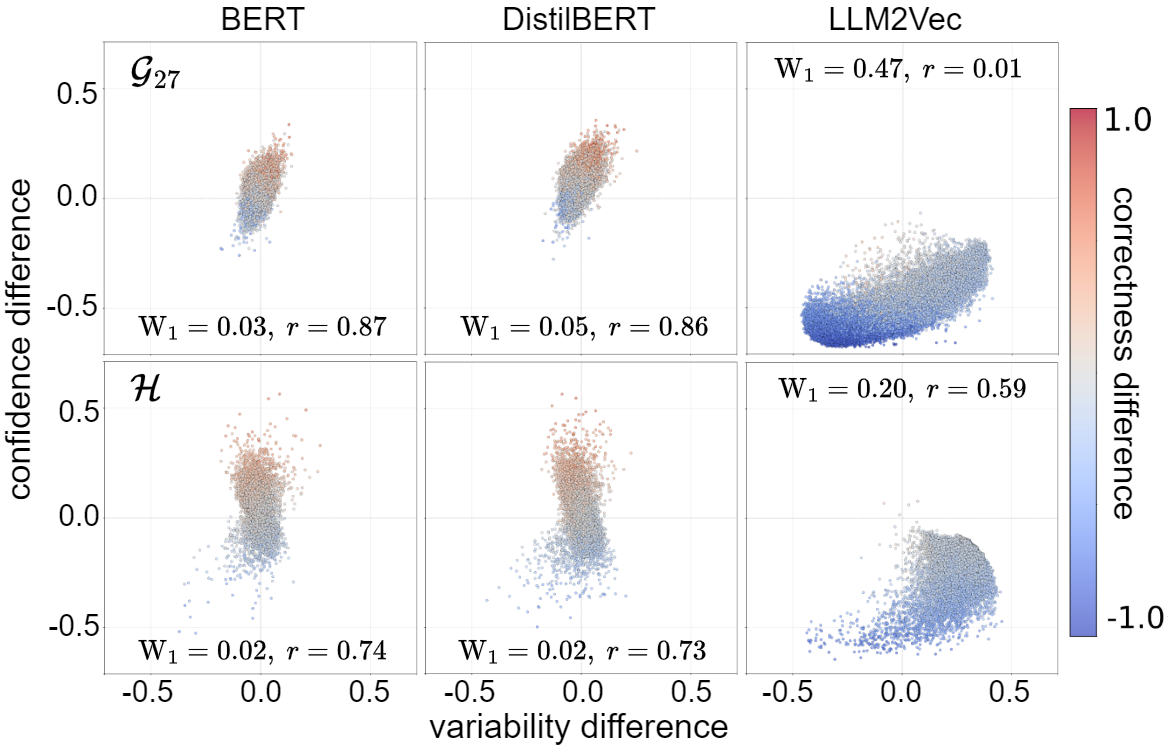}
    \caption{\label{fig:dep-encoder-comparison}Differences on the training dynamics across encoders with synthetic (top) and human (bottom) data on dependency parsing. Same notation as in Figure \ref{fig:sst5-encoder-comparison}.}
\end{figure}

\begin{figure}[h!]\centering
    \includegraphics[width=\linewidth]{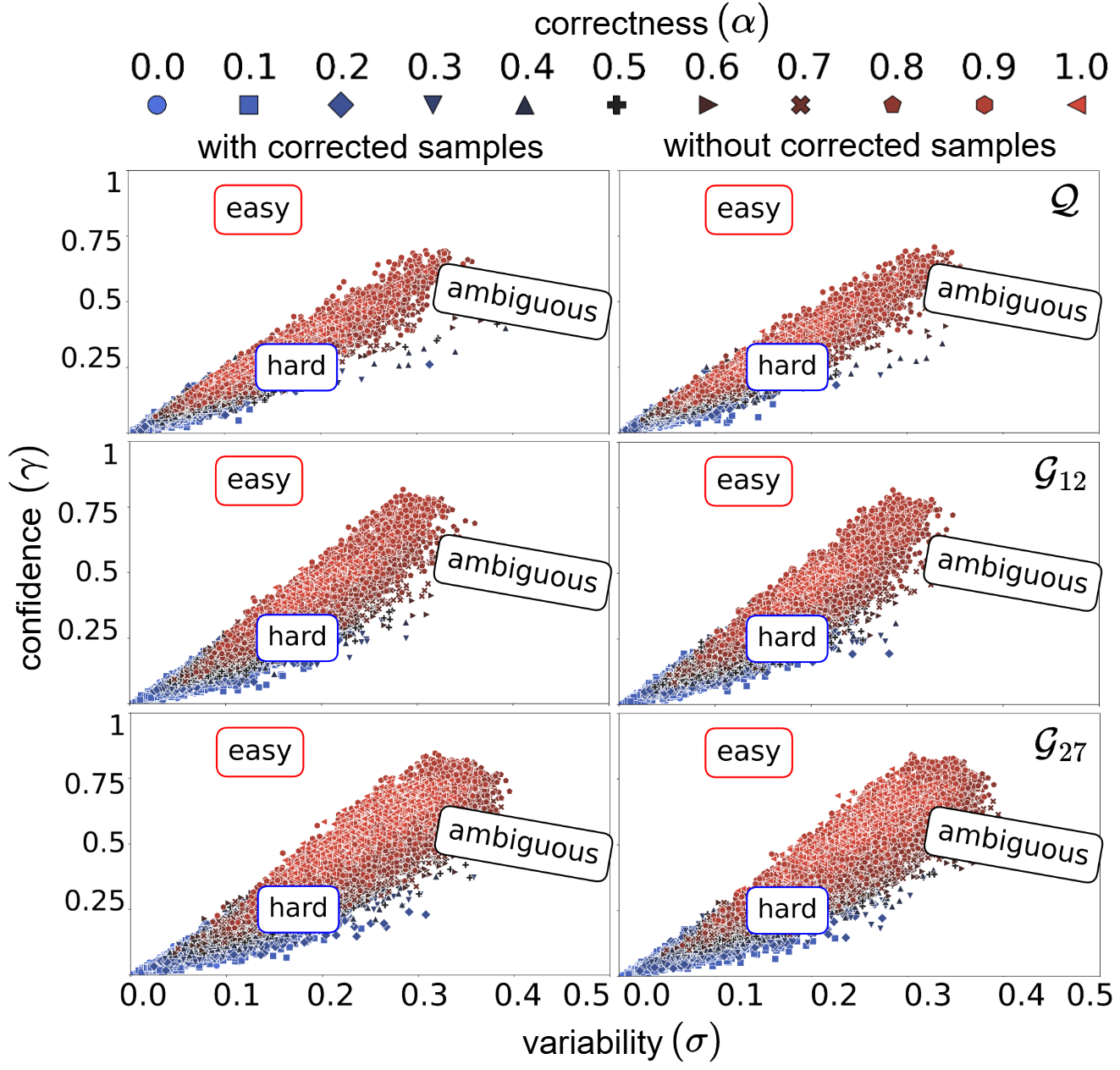}
    \caption{\label{fig:no_corrections} Dataset cartographies on dependency parsing task with LLM-generated English samples with the \textit{Difficulty-aware} prompt and RoBERTa encoder. The first column corresponds to the synthetic datasets after applying the corrections, whereas the second column contains only the samples that originally preserved the tree structure. Both distributions were subsampled to 20{,}000 instances to facilitate visualization. Same notation as in Figure \ref{fig:heatmap-comparison}.}
\end{figure}

\begin{figure*}[hptb!]\centering
    \includegraphics[width=\linewidth]{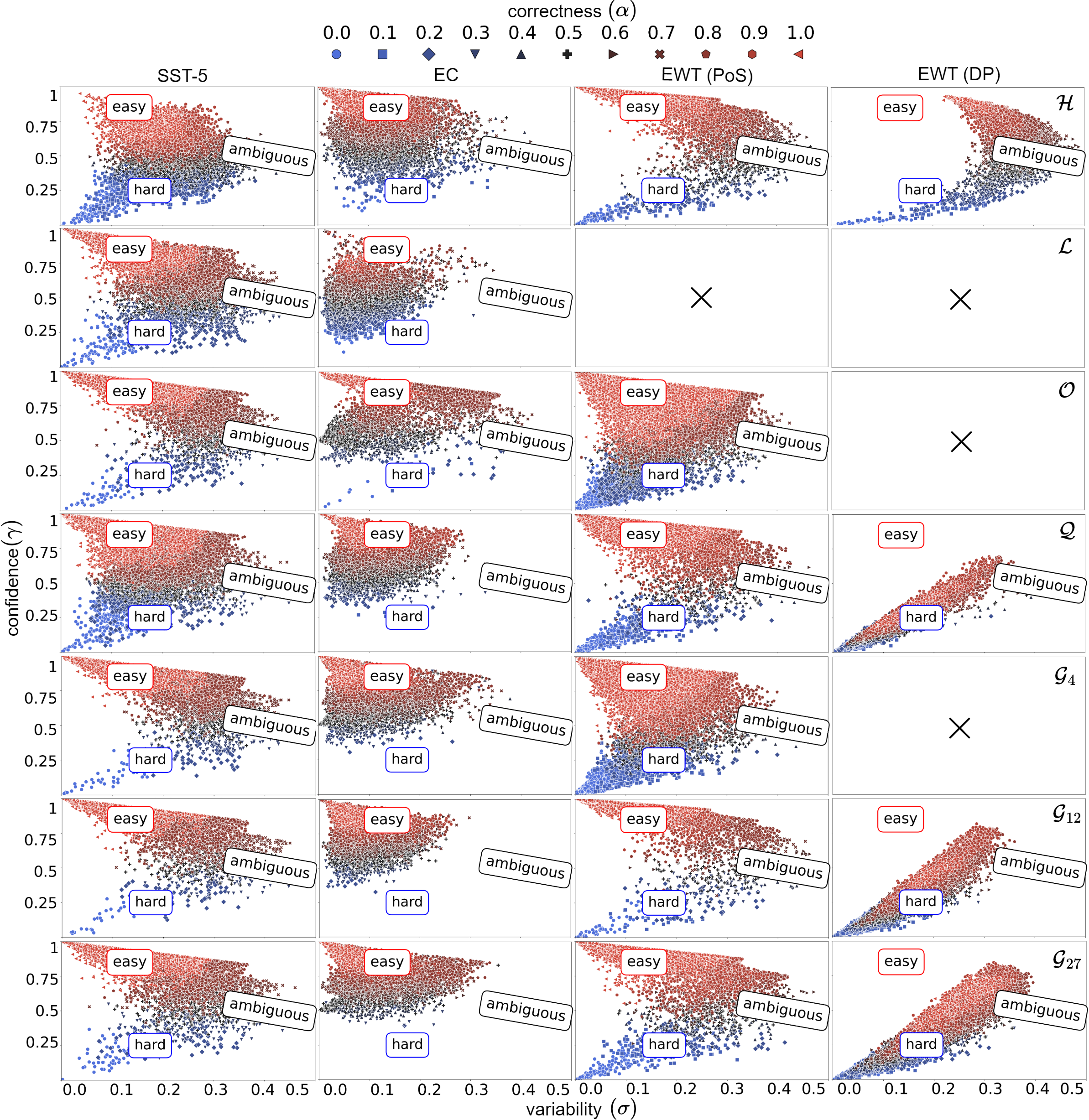}
    \caption{\label{fig:total} Dataset cartographies on all tasks with human and LLM-generated English samples with the \textit{Difficulty-aware} prompt and RoBERTa encoder. Same notation as in Figure \ref{fig:heatmap-comparison}.}
\end{figure*}

Figure \ref{fig:total} shows the human and synthetic-based dataset cartographies on English datasets. For the dependency parsing task, Figure \ref{fig:no_corrections} shows the data maps containing only the samples that originally preserved the tree structure and did not require any corrections. The nearly identical plots confirm that the observed training dynamics reflect the data characteristics rather than artifacts introduced by the correction process. Encoder comparisons are included in Figures \ref{fig:ec-encoder-comparison}-\ref{fig:dep-encoder-comparison} for English datasets.

The distributional comparison between prompts in terms of Wasserstein distance (\wasserstein) and Cliff's delta (\cliffdelta) is provided in Table \ref{tab:prompt_cliffs_english}. The results using the full datasets (human-authored and synthetic) are included in Table \ref{tab:performance-english}.

\section{Multilingual benchmark}\label{ap:multilingual-benchmark}
To extend our framework to a multilingual setting, we incorporate Spanish and French experiments. To establish a comparison of training dynamics across languages, we opted to translate the English SST-5 dataset using the NLLB-200 model \cite{nllbteam2022languageleftbehindscaling}, thereby maintaining parallel semantic structures. For the multi-label task, we used the Spanish version of the EC dataset (no French version available). For token-level and tree prediction tasks (PoS tagging and dependency parsing), we used UD treebanks: Spanish GSD and French Sequoia. 
Since text generation is time-consuming and generating enough data to reach the desired number of filtered instances required a large overgeneration factor, we subsampled the Spanish GSD corpus to 2{,}000 instances. French Sequoia already contains a relatively small number of samples, but we did not find Spanish UD datasets of comparable size. Therefore, all results reported for Spanish GSD are based on this 2{,}000 sample subset. We controlled translation quality by back-translating the data and measuring BLEU, ROUGE-L, TER, METEOR, and COMET \cite{rei-etal-2020-comet} scores (Table \ref{tab:translation-quality-sst5}). We relied on XLM-RoBERTa \cite{conneau2020unsupervisedcrosslingualrepresentationlearning} as the encoder to obtain the training dynamics in our analysis.

\begin{table}[htpb]\centering\footnotesize
    \setlength{\tabcolsep}{3pt}
    \begin{tabular}{c|ccccc}
    \hline 
     & BLEU & ROUGE & TER & METEOR & COMET \\
     \hline 
     Spanish & 43.2 & 0.77 & 29.8 & 0.77 & 0.83 \\
     French & 38.0 & 0.72 &  35.1 & 0.73 & 0.82\\
     \hline 
    \end{tabular}
    \caption{Translation quality for SST-5 data.}
    \label{tab:translation-quality-sst5}
\end{table}

\begin{figure}[htpb]\centering
    \includegraphics[width=\linewidth]{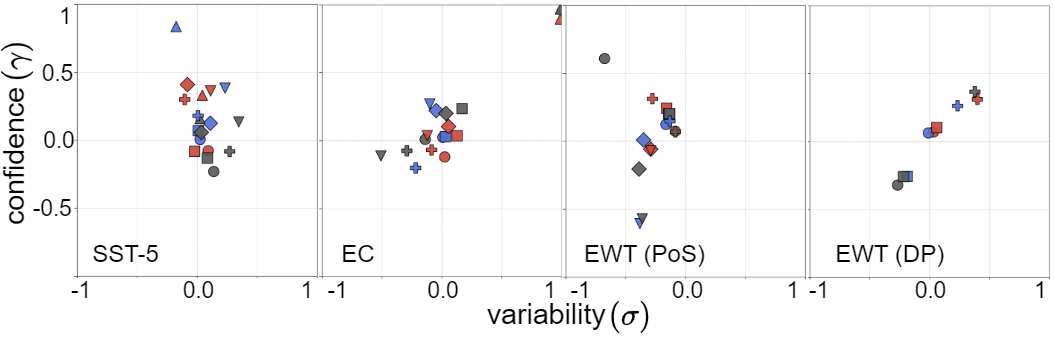}
    \caption{\label{fig:spanish-prompt-comparison}Comparison of training dynamics across prompting strategies on Spanish datasets. Subfigures illustrate the Cliff's delta for confidence ($\gamma$)  and variability ($\sigma$) across SST-5, EC, EWT (PoS) and EWT (DP) relatively to the \textit{naive} prompt. Same notation as in Figure \ref{fig:sst5-prompt-comparison}.}
\end{figure}

\begin{figure}[h!]\centering
    \includegraphics[width=\linewidth]{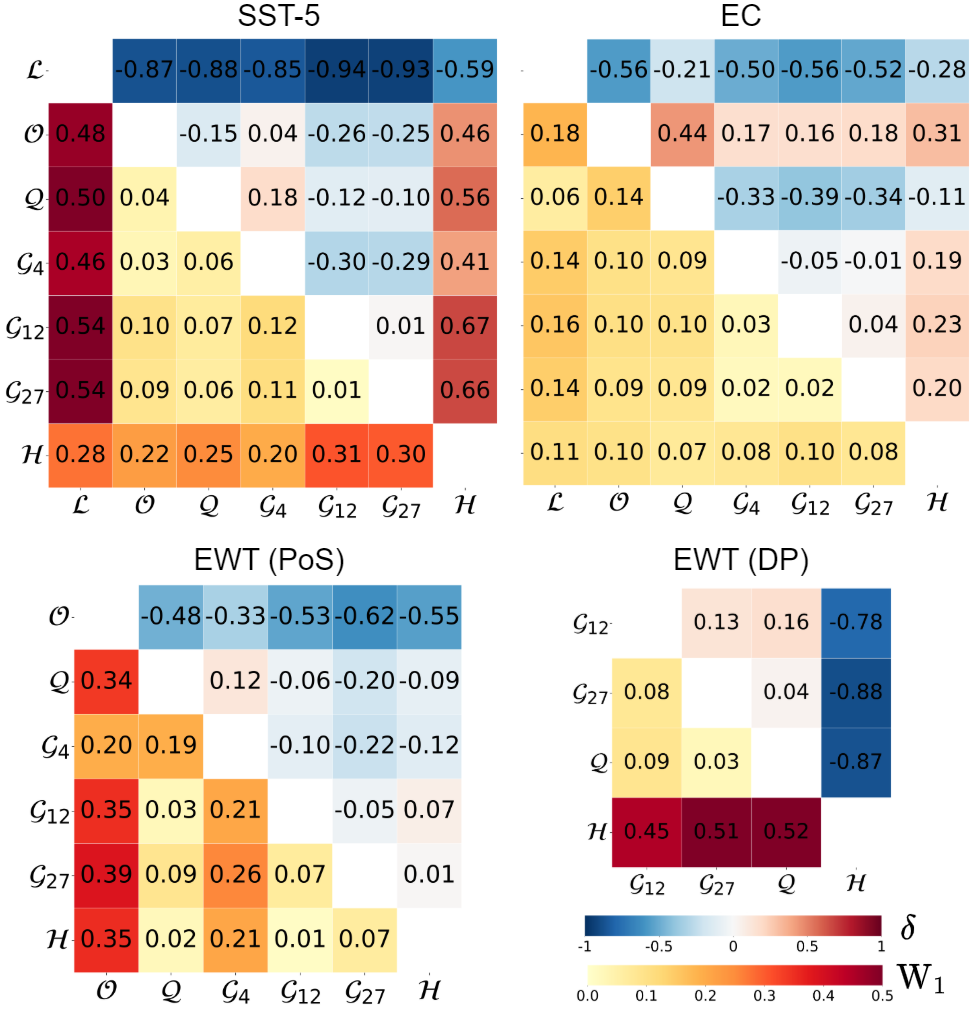}
    \caption{\label{fig:heatmap-comparison-spanish}
    Comparison of human and synthetic Spanish distributions across LLMs
    and the \textit{Difficulty-aware} prompt using Wasserstein distance on $\mathcal{T}^\theta_*$ (lower triangular matrices), and Cliff's delta on correctness (upper triangular matrices). Same notation as in Figure \ref{fig:llm-comparison-sst5}.}
\end{figure}

\begin{figure}[h!]\centering
    \includegraphics[width=\linewidth]{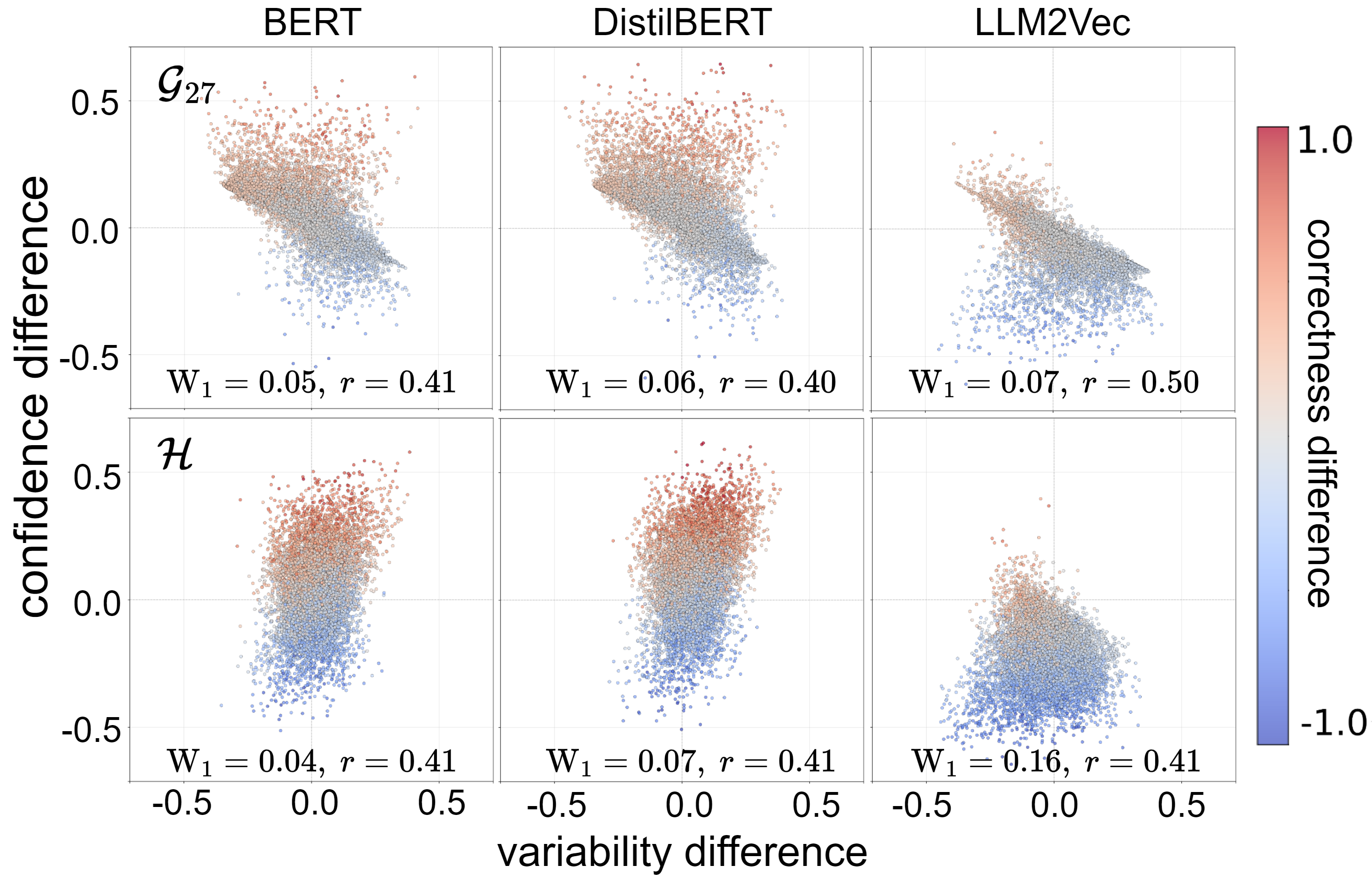}
    \caption{\label{fig:sst5s-encoder-comparison}Differences on the training dynamics across encoders with synthetic (top) and human (bottom) data on Spanish SST-5. Same notation as in Figure \ref{fig:sst5-encoder-comparison}.}
\end{figure}

\begin{figure*}[hptb]\centering
    \includegraphics[width=\linewidth]{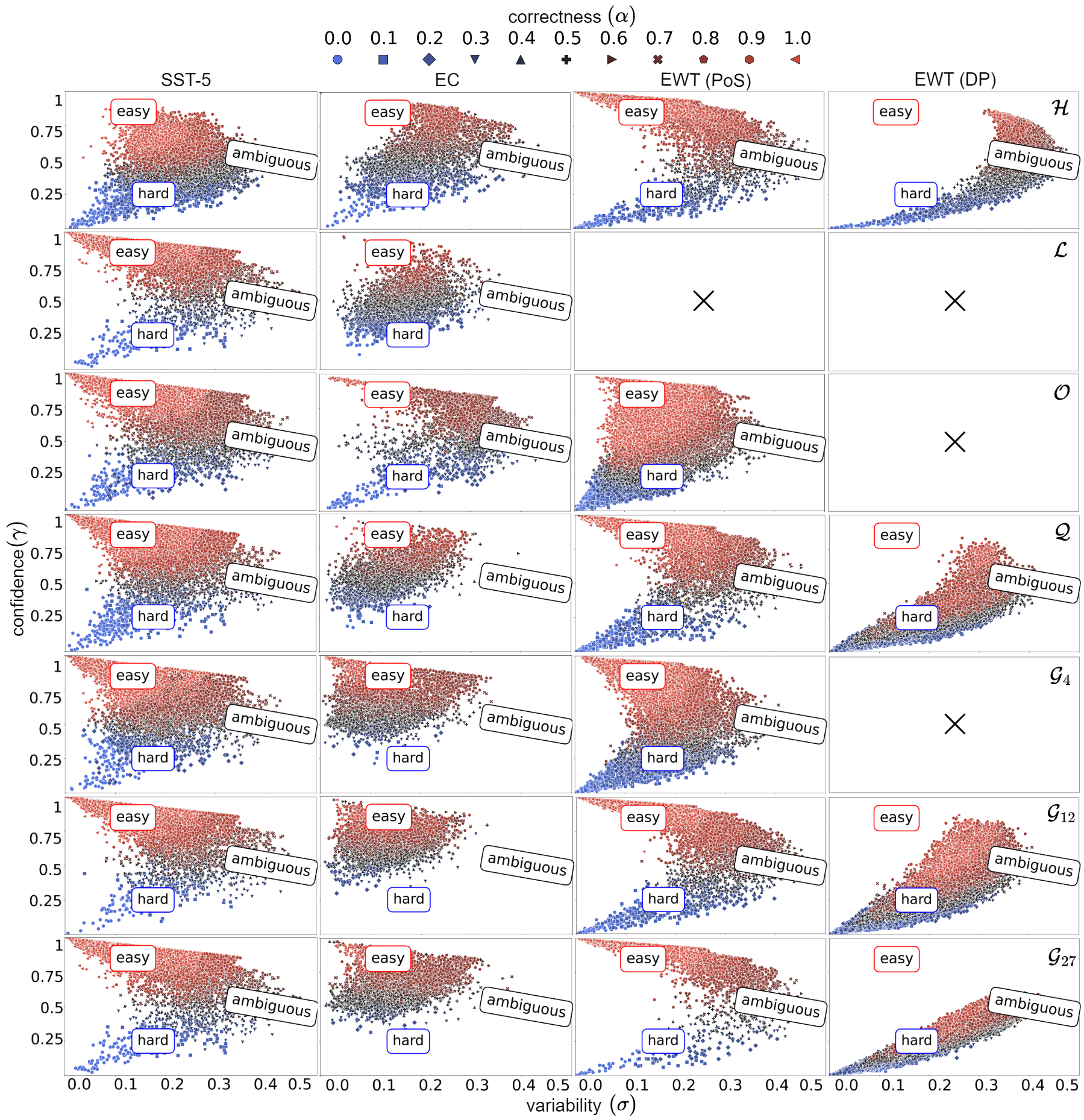}
    \caption{\label{fig:total_spanish} Dataset cartographies on all tasks with human and LLM-generated Spanish samples with the \textit{Difficulty-aware} prompt and RoBERTa encoder. Same notation as in Figure \ref{fig:heatmap-comparison}.}
\end{figure*}

\paragraph{Results on Spanish} Figures \ref{fig:spanish-prompt-comparison}, \ref{fig:heatmap-comparison-spanish}, \ref{fig:sst5s-encoder-comparison}, and \ref{fig:total_spanish} show the distributional comparison between prompts, the distributional comparison between LLMs, the encoder differences, and the dataset cartographies for Spanish human and synthetic datasets.

\begin{figure}[h!]\centering
    \includegraphics[width=\linewidth]{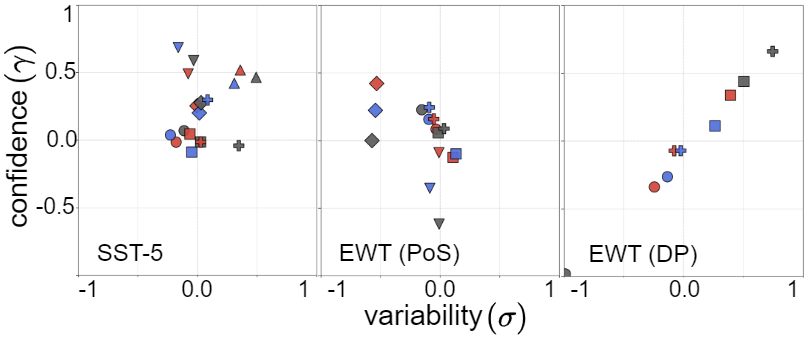}
    \caption{\label{fig:french-prompt-comparison}Comparison of training dynamics across prompting strategies on French datasets. Subfigures illustrate the Cliff's delta for confidence ($\gamma$)  and variability ($\sigma$) across SST-5, EC, EWT (PoS) and EWT (DP) relatively to the \textit{naive} prompt. Same notation as in Figure \ref{fig:sst5-prompt-comparison}.}
\end{figure}

\begin{figure}[h!]\centering
    \includegraphics[width=\linewidth]{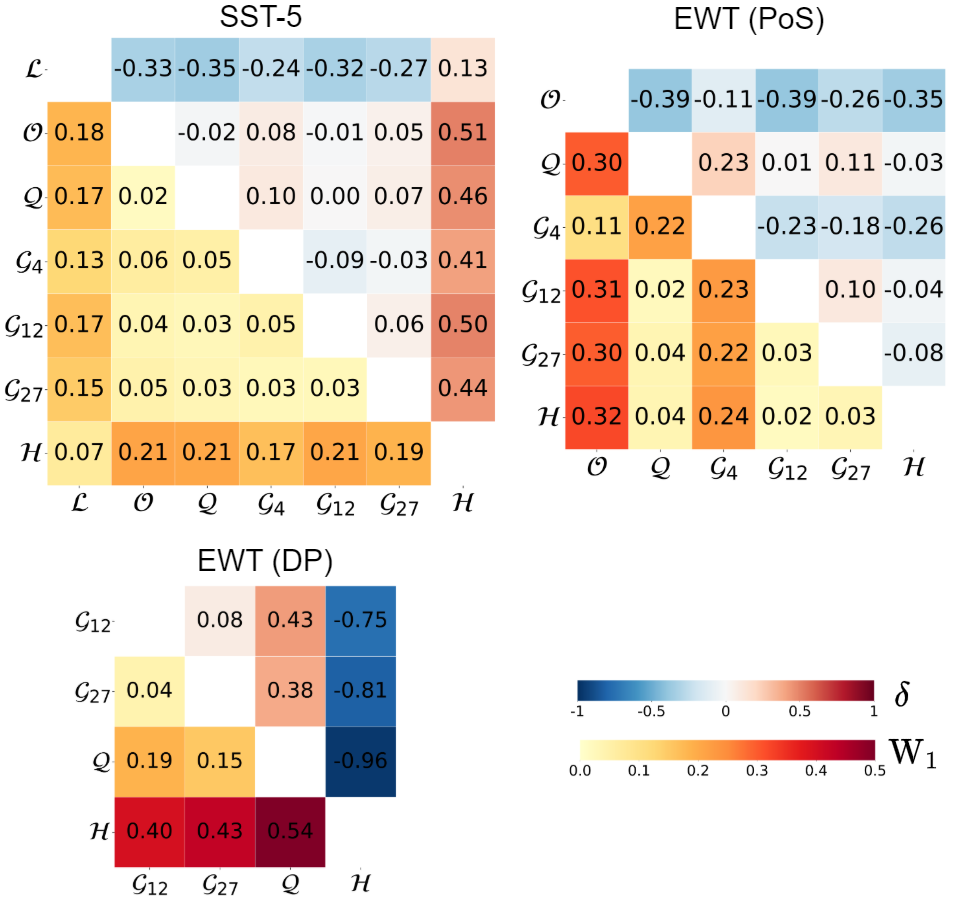}
    \caption{\label{fig:heatmap-comparison-french}
    Comparison of human and synthetic French distributions across LLMs
    and the \textit{Difficulty-aware} prompt using Wasserstein distance on $\mathcal{T}^\theta_*$ (lower triangular matrices), and Cliff's delta on correctness (upper triangular matrices). Same notation as in Figure \ref{fig:llm-comparison-sst5}.}
\end{figure}

\begin{figure}[h!]\centering
    \includegraphics[width=\linewidth]{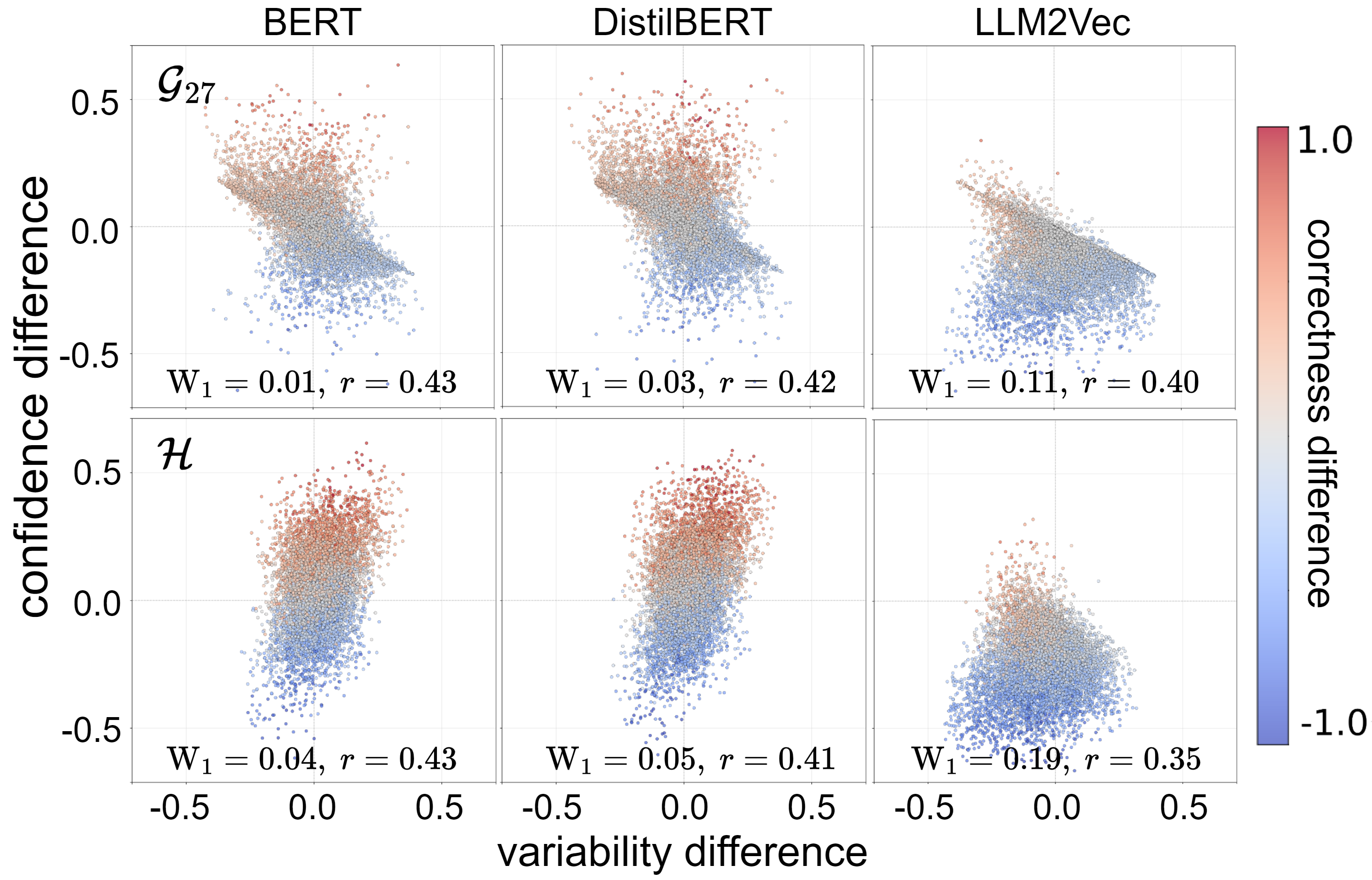}
    \caption{\label{fig:sst5f-encoder-comparison}Differences on the training dynamics across encoders with synthetic (top) and human (bottom) data on French SST-5. Same notation as in Figure \ref{fig:sst5-encoder-comparison}.}
\end{figure}

\begin{figure*}[hptb]\centering
    \includegraphics[width=0.8\linewidth]{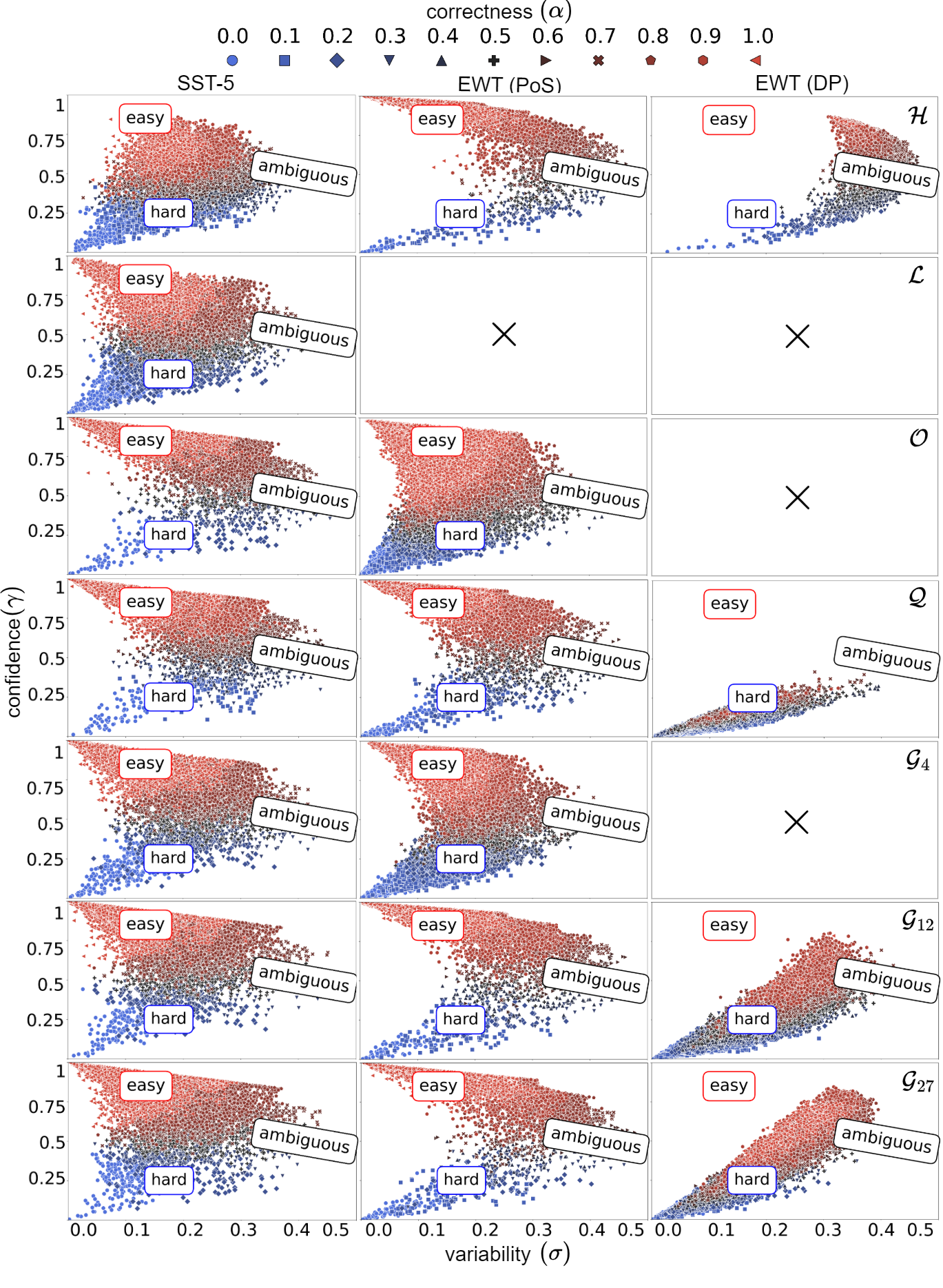}
    \caption{\label{fig:total_french} Dataset cartographies on all tasks with human and LLM-generated French samples with the \textit{Difficulty-aware} prompt and RoBERTa encoder. Same notation as in Figure \ref{fig:heatmap-comparison}.}
\end{figure*}

\paragraph{Results on French} As in the Spanish case, Figures \ref{fig:french-prompt-comparison}, \ref{fig:heatmap-comparison-french}, \ref{fig:sst5f-encoder-comparison} and \ref{fig:total_french} present the corresponding analyses for French datasets.

In both languages, differences between prompts exhibit a broader range than in the English setting. There are even extreme cases, such as the point near $(1,1)$ in Figure \ref{fig:spanish-prompt-comparison} for the multi-label task, indicating that the \textit{exemplified} and \textit{Difficulty-aware} prompts on the LLaMA model differ substantially from the \textit{naive} prompt, showing higher confidence and variability in their distributions. This is consistent with the test results (Table \ref{tab:performance-multilingual}), where the data generated by LLaMA for the multi-label task appears to degrade substantially when no examples are provided in the prompt.

Tables \ref{tab:performance-multilingual} and \ref{fig:subset-multilingual} show the performance summary of the multilingual benchmark. For OOD evaluation, we use the same datasets as in the English experiments: the Yelp dataset translated with NLLB-200 and the UD PUD dataset corresponding to each language.

\begin{table*}[h!]
    \scriptsize\centering
    \setlength{\tabcolsep}{2.5pt}
    \begin{tabular}{ll|cccc|cccc|cccc|cccc}
        \hline
         & & \multicolumn{4}{c|}{{SST-5}}
         & \multicolumn{4}{c|}{{EC}} 
         & \multicolumn{4}{c|}{EWT\textsubscript{PoS}} 
         & \multicolumn{4}{c}{{EWT\textsubscript{DP}}} \\
         
         & & \wasserstein & $\alpha$ & $\sigma$ & $\gamma$
         & \wasserstein & $\alpha$ & $\sigma$ & $\gamma$
         & \wasserstein & $\alpha$ & $\sigma$ & $\gamma$
         & \wasserstein & $\alpha$ & $\sigma$ & $\gamma$ \\
        \hline

        \parbox[t]{2mm}{\multirow{6}{*}{\rotatebox[origin=c]{90}{\textit{exemp.}}}}
         & \llama & 0.11 & 0.22 & 0.06 & 0.39 
                  & 0.03 & -0.08 & -0.13 & -0.02
                  & \nan & \nan & \nan & \nan
                  & \nan & \nan & \nan & \nan \\
         & \olmo & 0.09 & 0.18 & -0.08 & 0.32
                 & 0.10 & -0.26 & 0.12 & -0.25
                 & 0.18 & 0.07 & 0.13 & 0.02
                 & \nan & \nan & \nan & \nan \\
         & \qwen & 0.03 & 0.03 & 0.11 & 0.06
                & 0.03 & 0.10 & -0.01 & 0.12
                & 0.07 & 0.04 & 0.01 & 0.07
                & 0.03 & 0.09 & -0.06 & 0.11 \\
         & \gemma{4} & 0.06 & 0.12 & -0.08 & 0.21
                     & 0.03 & 0.11 & 0.04 & 0.12
                     & 0.03 & -0.11 & -0.03 & -0.17
                     & \nan & \nan & \nan & \nan \\
         & \gemma{12} & 0.01 & -0.02 & 0.05 & -0.06
                      & 0.01 & 0.06 & 0.04 & 0.07
                      & 0.10 & -0.06 & 0.07 & -0.12
                      & 0.08 & 0.13 & -0.20 & 0.32 \\
         & \gemma{27} & 0.02 & 0.05 & -0.13 & 0.06
                      & 0.02 & 0.05 & 0.00 & 0.07
                      & 0.03 & -0.16 & 0.06 & -0.28
                      & 0.10 & 0.19 & 0.00 & 0.35 \\
        \hline

        \parbox[t]{2mm}{\multirow{6}{*}{\rotatebox[origin=c]{90}{\textit{diff.}}}}
         & \llama & 0.10 & 0.18 & 0.20 & 0.36
                  & 0.07 & -0.23 & -0.33 & -0.17
                  & \nan & \nan & \nan & \nan
                  & \nan & \nan & \nan & \nan \\
         & \olmo & 0.10 & 0.17 & 0.01 & 0.32
                 & 0.03 & -0.08 & 0.07 & -0.10
                 & 0.16 & 0.00 & 0.08 & -0.07
                 & \nan & \nan & \nan & \nan \\
         & \qwen & 0.03 & -0.11 & 0.07 & -0.10
                & 0.02 & -0.05 & -0.11 & -0.05
                & 0.09 & 0.11 & -0.05 & 0.15
                & 0.01 & -0.03 & 0.03 & -0.07 \\
         & \gemma{4} & 0.04 & 0.09 & -0.05 & 0.15
                     & 0.02 & 0.06 & -0.09 & 0.09
                     & 0.04 & -0.12 & -0.03 & -0.20
                     & \nan & \nan & \nan & \nan \\
         & \gemma{12} & 0.01 & 0.00 & -0.01 & 0.00
                      & 0.03 & 0.10 & 0.05 & 0.14
                      & 0.27 & 0.26 & 0.05 & 0.32
                      & 0.13 & 0.23 & -0.17 & 0.46 \\
         & \gemma{27} & 0.01 & -0.01 & 0.02 & -0.05
                      & 0.01 & 0.06 & -0.04 & 0.07
                      & 0.06 & 0.01 & -0.04 & -0.01
                      & 0.14 & 0.29 & 0.09 & 0.45 \\
        \hline

        \parbox[t]{2mm}{\multirow{6}{*}{\rotatebox[origin=c]{90}{\textit{ambig.}}}}
         & \llama & 0.08 & 0.15 & 0.17 & 0.26
                  & 0.05 & -0.08 & -0.24 & -0.02
                  & \nan & \nan & \nan & \nan
                  & \nan & \nan & \nan & \nan \\
         & \olmo & 0.09 & 0.13 & 0.12 & 0.28
                 & 0.18 & -0.53 & 0.20 & -0.50
                 & 0.12 & -0.09 & 0.08 & -0.17
                 & \nan & \nan & \nan & \nan \\
         & \qwen & 0.03 & -0.07 & 0.20 & -0.02
                & 0.03 & 0.04 & 0.10 & 0.05
                & 0.08 & 0.11 & -0.09 & 0.17
                & 0.02 & 0.04 & -0.16 & -0.01 \\
         & \gemma{4} & 0.02 & 0.06 & 0.10 & 0.04
                     & 0.04 & 0.14 & -0.14 & 0.19
                     & 0.08 & -0.14 & -0.02 & -0.25
                     & \nan & \nan & \nan & \nan \\
         & \gemma{12} & 0.03 & 0.05 & -0.08 & 0.12
                      & 0.01 & -0.02 & 0.02 & -0.04
                      & 0.33 & 0.30 & 0.09 & 0.35
                      & 0.06 & 0.09 & -0.05 & 0.15 \\
         & \gemma{27} & 0.04 & -0.10 & 0.24 & -0.17
                      & 0.04 & 0.14 & 0.04 & 0.16
                      & 0.01 & 0.01 & -0.08 & -0.00
                      & 0.15 & 0.30 & 0.02 & 0.50 \\
        \hline
    \end{tabular}
    \caption{Comparison of prompt strategies with respect to the naive prompt using the Wasserstein distance (\wasserstein) and Cliff's delta ($\delta$) over confidence ($\gamma$), variability ($\sigma$) and correctness ($\alpha$) across the English datasets. Same notation as Figure \ref{fig:heatmap-comparison} for LLM symbols. The symbol (\nan) denotes runs where the LLM could not produce valid formatted or meaningful samples.}
    \label{tab:prompt_cliffs_english}
\end{table*}

\begin{table*}[h!]
    \scriptsize\centering
    \setlength{\tabcolsep}{2.5pt}
    \begin{tabular}{ll|cccc|ccc}
        \hline
            && \multicolumn{4}{c|}{\textit{in-distribution}} & \multicolumn{3}{c}{\textit{out-of-distribution}}\\
         & & {\tiny SST-5} & {\tiny EC} & {\tiny EWT\textsubscript{PoS}} & {\tiny EWT\textsubscript{DP}} & {\tiny SST-5} & {\tiny EWT\textsubscript{PoS}} & {\tiny EWT\textsubscript{DP}} \\
         \hline

         \parbox[t]{2mm}{\multirow{6}{*}{\rotatebox[origin=c]{90}{\textit{naive}}}} & \llama & 40.5 & 30.9 & \nan & \nan & 49.2 & \nan & \nan \\
         & \olmo & 35.6 & 27.4 & 75.0 & \nan & 46.1 & 77.9 & \nan \\
         & \qwen & 37.5 & 39.7 & 81.1 & 12.1 & 46.0 & 86.2 & 9.8 \\
         & \gemma{4} & 39.3 & 33.7 & 69.5 & \nan & 44.4 & 73.0 & \nan \\
         & \gemma{12} & 42.2 & 35.3 & 81.0 & 12.7 & 47.6 & 86.4 & 12.0 \\
         & \gemma{27} & 42.5 & 41.4 & 83.7 & 14.6 & 48.7 & 88.6 & 15.3 \\
         \hline

         \parbox[t]{2mm}{\multirow{6}{*}{\rotatebox[origin=c]{90}{\textit{exemp.}}}} & \llama & 35.0 & 31.7 & \nan & \nan & 48.3 & \nan & \nan \\
         & \olmo & 40.9 & 29.6 & 75.5 & \nan & 30.4 & 79.3 & \nan \\
         & \qwen & 40.1 & 37.6 & 81.9 & 10.5 & 45.2 & 86.8 & 8.7 \\
         & \gemma{4} & 28.6 & 37.2 & 74.0 & \nan & 47.3 & 77.9 & \nan \\
         & \gemma{12} & 39.8 & 38.2 & 82.1 & 21.3 & 43.3 & 87.0 & 24.3 \\
         & \gemma{27} & 19.2 & 43.5 & 83.8 & 36.2 & 45.9 & 88.8 & 39.5 \\
         \hline

         \parbox[t]{2mm}{\multirow{6}{*}{\rotatebox[origin=c]{90}{\textit{diff.}}}} & \llama & 40.7 & 38.6 & \nan & \nan & 38.9 & \nan & \nan \\
         & \olmo & 43.0 & 26.4 & 71.9 & \nan & 37.6 & 74.9 & \nan \\
         & \qwen & 40.4 & 40.8 & 82.2 & 8.5 & 38.7 & 86.7 & 8.6 \\
         & \gemma{4} & 37.2 & 36.8 & 73.1 & \nan & 46.6 & 77.3 & \nan \\
         & \gemma{12} & 38.3 & 38.4 & 80.5 & 24.9 & 44.1 & 86.6 & 27.9 \\
         & \gemma{27} & 41.6 & 43.2 & 83.9 & 35.9 & 53.9 & 88.8 & 38.9 \\
         \hline

         \parbox[t]{2mm}{\multirow{6}{*}{\rotatebox[origin=c]{90}{\textit{amb.}}}} & \llama & 44.1 & 38.6 & \nan & \nan & 48.6 & \nan & \nan \\
         & \olmo & 40.3 & 34.2 & 72.4 & \nan & 44.8 & 75.9 & \nan \\
         & \qwen & 41.4 & 40.5 & 82.3 & 8.6 & 28.3 & 86.7 & 7.3 \\
         & \gemma{4} & 40.8 & 37.7 & 71.0 & \nan & 49.0 & 75.8 & \nan \\
         & \gemma{12} & 30.5 & 40.1 & 81.1 & 23.8 & 47.6 & 87.0 & 25.9 \\
         & \gemma{27} & 40.6 & 42.3 & 83.9 & 35.6 & 44.6 & 88.5 & 37.1 \\
         \hline

         \multicolumn{2}{c|}{$\mathcal{H}$} & 58.1 & 59.0 & 93.0 & 93.2 & 57.1 & 94.4 & 92.3 \\
         \hline
    \end{tabular}
    \caption{\label{tab:performance-english}Performance of RoBERTa on synthetic data generated from different prompts, LLMs and English datasets (SST-5 for sentiment analysis, EC for multi-label classification and EWT for PoS-tagging and dependency parsing). The symbol (\nan) denotes runs where the LLM could not produce valid formatted or meaningful samples. Last row ($\mathcal{H}$) denotes the performance with the full original dataset.}
\end{table*}

\clearpage


\begin{table*}[h!]
    \scriptsize\centering
    \setlength{\tabcolsep}{2.5pt}
    \begin{tabular}{l l|cc|c|cc|cc|cc|cc|cc}
        \hline
            && \multicolumn{7}{c|}{\textit{in-distribution}} & \multicolumn{6}{c}{\textit{out-of-distribution}} \\
         & & \multicolumn{2}{c|}{\tiny SST-5} & \multicolumn{1}{c|}{\tiny EC}
         & \multicolumn{2}{c|}{\tiny EWT\textsubscript{PoS}}
         & \multicolumn{2}{c|}{\tiny EWT\textsubscript{DP}}
         & \multicolumn{2}{c|}{\tiny SST-5}
         & \multicolumn{2}{c|}{\tiny EWT\textsubscript{PoS}}
         & \multicolumn{2}{c}{\tiny EWT\textsubscript{DP}} \\
         & & \tiny es & \tiny fr & \tiny es & \tiny es & \tiny fr & \tiny es & \tiny fr 
         & \tiny es & \tiny fr & \tiny es & \tiny fr & \tiny es & \tiny fr \\
        \hline
     
        % ===================== NAIVE =====================
        \parbox[t]{2mm}{\multirow{6}{*}{\rotatebox[origin=c]{90}{\textit{naive}}}}
        
        & \llama & 37.6 & 35.4 & 4.9 & \nan & \nan & \nan & \nan & 35.3 & 42.6 & \nan & \nan & \nan & \nan \\
        & \olmo  & 34.4 & 36.2 & 24.2 & 75.5 & 68.5 & \nan & \nan & 45.9 & 36.2 & 79.5 & 71.0 & \nan & \nan \\
        & \qwen  & 36.3 & 37.6 & 36.0 & 87.5 & 84.9 & 7.2 & 7.3 & 42.8 & 46.4 & 89.5 & 87.7 & 9.5 & 6.8 \\
        & \gemma{4}  & 32.1 & 32.9 & 30.9 & 77.6 & 77.9 & \nan & \nan & 43.2 & 44.5 & 80.3 & 80.4 & \nan & \nan \\
        & \gemma{12} & 39.5 & 38.6 & 29.1 & 86.8 & 86.9 & 6.3 & 5.3 & 45.4 & 42.4 & 90.3 & 88.6 & 8.1 & 6.0 \\
        & \gemma{27} & 39.7 & 37.9 & 39.0 & 86.9 & 87.3 & 1.8 & 10.3 & 49.3 & 46.8 & 89.8 & 89.5 & 2.5 & 10.6 \\
        
        \hline
        
        % ===================== EXEMP =====================
        \parbox[t]{2mm}{\multirow{6}{*}{\rotatebox[origin=c]{90}{\textit{exemp.}}}}
        
        & \llama & 34.5 & 35.5 & 32.3 & \nan & \nan & \nan & \nan & 38.3 & 38.6 & \nan & \nan & \nan & \nan \\
        & \olmo  & 38.1 & 38.5 & 27.1 & 69.4 & 65.5 & \nan & \nan & 36.3 & 41.8 & 77.6 & 68.0 & \nan & \nan \\
        & \qwen  & 40.5 & 36.1 & 31.9 & 87.5 & 86.8 & 6.9 & 7.1 & 42.9 & 37.8 & 88.0 & 89.1 & 8.9 & 6.6 \\
        & \gemma{4}  & 35.5 & 31.0 & 35.6 & 75.0 & 77.7 & \nan & \nan & 42.7 & 44.3 & 77.3 & 79.0 & \nan & \nan \\
        & \gemma{12} & 38.4 & 37.0 & 34.5 & 88.9 & 86.0 & 12.9 & 13.0 & 43.0 & 44.7 & 89.4 & 88.3 & 15.5 & 14.2 \\
        & \gemma{27} & 39.5 & 36.5 & 39.3 & 88.8 & 88.2 & 1.3 & 27.7 & 45.9 & 40.4 & 90.7 & 90.0 & 1.9 & 31.4 \\
        
        \hline
        
        % ===================== DIFF =====================
        \parbox[t]{2mm}{\multirow{6}{*}{\rotatebox[origin=c]{90}{\textit{diff.}}}}
        & \llama & 23.1 & 35.7 & 32.6 & \nan & \nan & \nan & \nan & 20.0 & 39.7 & \nan & \nan & \nan & \nan \\
        & \olmo  & 38.7 & 39.4 & 23.5 & 57.7 & 56.8 & \nan & \nan & 41.4 & 40.1 & 58.2 & 58.8 & \nan & \nan \\
        & \qwen  & 37.2 & 38.9 & 38.1 & 87.8 & 85.8 & 8.2 & 8.3 & 45.0 & 45.8 & 88.9 & 87.9 & 10.6 & 8.0 \\
        & \gemma{4}  & 36.4 & 35.9 & 33.9 & 75.5 & 72.3 & \nan & \nan & 42.3 & 40.7 & 78.8 & 76.3 & \nan & \nan \\
        & \gemma{12} & 40.2 & 37.5 & 35.0 & 87.6 & 88.3 & 23.7 & 28.3 & 46.0 & 44.5 & 89.0 & 89.5 & 26.9 & 30.5 \\
        & \gemma{27} & 37.4 & 35.7 & 41.4 & 88.8 & 87.9 & 27.8 & 22.4 & 46.7 & 48.4 & 89.3 & 89.5 & 33.5 & 24.8 \\
        
        \hline
        
        % ===================== AMB =====================
        \parbox[t]{2mm}{\multirow{6}{*}{\rotatebox[origin=c]{90}{\textit{amb.}}}}
        & \llama & 35.1 & 36.8 & 32.7 & \nan & \nan & \nan & \nan & 40.1 & 42.7 & \nan & \nan & \nan & \nan \\
        & \olmo  & 35.0 & 32.2 & 28.6 & 41.7 & 53.5 & \nan & \nan & 40.0 & 33.7 & 41.6 & 53.5 & \nan & \nan \\
        & \qwen  & 40.3 & 38.7 & 35.9 & 87.8 & 86.6 & 6.6 & 22.1 & 42.5 & 42.6 & 89.0 & 87.7 & 8.2 & 25.2 \\
        & \gemma{4}  & 29.8 & 36.0 & 33.6 & 66.6 & 67.9 & \nan & \nan & 43.7 & 43.9 & 70.6 & 69.4 & \nan & \nan \\
        & \gemma{12} & 41.2 & 43.1 & 33.6 & 88.9 & 86.6 & 23.5 & 23.8 & 45.6 & 44.5 & 89.7 & 87.5 & 27.5 & 26.3 \\
        & \gemma{27} & 36.7 & 35.7 & 37.8 & 89.3 & 87.3 & 27.9 & 23.3 & 46.6 & 48.3 & 89.9 & 88.6 & 32.5 & 27.0 \\
        
        \hline

         \multicolumn{2}{c|}{$\mathcal{H}$} & 50.8 & 51.8 & 59.0 & 94.9 & 97.2 & 90.3 & 94.9 & 49.7 & 48.0 & 92.0 & 95.5 & 82.1 & 84.6 \\
        \hline
    \end{tabular}
    \caption{\label{tab:performance-multilingual}Performance of XLM-RoBERTa on synthetic data, generated with different prompts and LLMs in our multilingual benchmark. Same notation as in Table \ref{tab:performance-english}.}
\end{table*}

\begin{table*}[ht!]
    \scriptsize\centering
    \setlength{\tabcolsep}{2.5pt}
    \begin{tabular}{ll|cc|c|cc|cc|cc|cc|cc}
        \hline
            && \multicolumn{7}{c|}{\textit{in-distribution}} & \multicolumn{6}{c}{\textit{out-of-distribution}}\\
         & & \multicolumn{2}{c|}{SST-5} & {EC} & \multicolumn{2}{c|}{EWT\textsubscript{PoS}} & \multicolumn{2}{c|}{EWT\textsubscript{DP}} & \multicolumn{2}{c|}{SST-5} & \multicolumn{2}{c|}{EWT\textsubscript{PoS}} & \multicolumn{2}{c}{EWT\textsubscript{DP}} \\
         & & {es} & {fr} & {es} & {es} & {fr} & {es} & {fr} & {es} & {fr} & {es} & {fr} & {es} & {fr} \\
         \hline
         \multicolumn{1}{l}{\parbox[t]{2mm}{\multirow{5}{*}{\rotatebox[origin=c]{90}{\textit{full}}}}} & $\mathcal{L}$ & \cellcolor[RGB]{234,176,163}{23.1} & \cellcolor[RGB]{235,227,160}{35.7} & \cellcolor[RGB]{236,227,159}{32.6} & \nan & \nan & \nan & \nan & \cellcolor[RGB]{232,168,168}{20.0} & \cellcolor[RGB]{199,218,165}{39.7} & \nan & \nan & \nan & \nan \\
         & $\mathcal{O}$ & \cellcolor[RGB]{218,222,162}{38.7} & \cellcolor[RGB]{211,221,163}{39.4} & \cellcolor[RGB]{237,199,160}{23.5} & \cellcolor[RGB]{238,218,159}{57.7} & \cellcolor[RGB]{236,191,160}{56.8} & \nan & \nan & \cellcolor[RGB]{189,215,167}{41.4} & \cellcolor[RGB]{196,217,166}{40.1} & \cellcolor[RGB]{238,218,159}{58.2} & \cellcolor[RGB]{237,199,160}{58.8} & \nan & \nan \\
         & $\mathcal{Q}$ & \cellcolor[RGB]{228,225,161}{37.2} & \cellcolor[RGB]{215,222,163}{38.9} & \cellcolor[RGB]{213,221,163}{38.1} & \cellcolor[RGB]{169,205,162}{87.8} & \cellcolor[RGB]{177,210,166}{85.8} & \cellcolor[RGB]{233,173,165}{8.2} & \cellcolor[RGB]{233,173,165}{8.3} & \cellcolor[RGB]{174,208,164}{45.0} & \cellcolor[RGB]{169,206,163}{45.8} & \cellcolor[RGB]{163,202,160}{88.9} & \cellcolor[RGB]{171,207,163}{87.9} & \cellcolor[RGB]{234,175,164}{10.6} & \cellcolor[RGB]{233,173,165}{8.0} \\
         & $\mathcal{G}_{12}$ & \cellcolor[RGB]{208,220,164}{40.2} & \cellcolor[RGB]{224,224,161}{37.5} & \cellcolor[RGB]{226,225,161}{35.0} & \cellcolor[RGB]{169,205,163}{87.6} & \cellcolor[RGB]{173,208,164}{88.3} & \cellcolor[RGB]{235,183,160}{23.7} & \cellcolor[RGB]{236,190,160}{28.3} & \cellcolor[RGB]{170,206,163}{46.0} & \cellcolor[RGB]{174,208,164}{44.5} & \cellcolor[RGB]{163,202,160}{89.0} & \cellcolor[RGB]{168,205,162}{89.5} & \cellcolor[RGB]{236,195,160}{26.9} & \cellcolor[RGB]{237,202,160}{30.5} \\
         & $\mathcal{H}$ & \cellcolor[RGB]{165,203,161}{50.8} & \cellcolor[RGB]{160,200,159}{51.8} & \cellcolor[RGB]{158,199,158}{59.0} & \cellcolor[RGB]{159,199,158}{94.9} & \cellcolor[RGB]{158,199,158}{97.2} & \cellcolor[RGB]{158,199,158}{90.3} & \cellcolor[RGB]{158,199,158}{94.9} & \cellcolor[RGB]{159,199,158}{49.7} & \cellcolor[RGB]{162,201,160}{48.0} & \cellcolor[RGB]{158,199,158}{92.0} & \cellcolor[RGB]{158,199,158}{95.5} & \cellcolor[RGB]{158,199,158}{82.1} & \cellcolor[RGB]{158,199,158}{84.6} \\
         \hline
         \multicolumn{1}{l}{\parbox[t]{2mm}{\multirow{5}{*}{\rotatebox[origin=c]{90}{\textit{random}}}}} & $\mathcal{L}$ & \cellcolor[RGB]{232,168,168}{17.6} & \cellcolor[RGB]{239,225,159}{34.5} & \cellcolor[RGB]{231,226,160}{33.9} & \nan & \nan & \nan & \nan & \cellcolor[RGB]{232,168,168}{20.0} & \cellcolor[RGB]{175,209,165}{44.1} & \nan & \nan & \nan & \nan \\
         & $\mathcal{O}$ & \cellcolor[RGB]{212,221,163}{39.6} & \cellcolor[RGB]{226,225,161}{37.1} & \cellcolor[RGB]{236,195,160}{22.4} & \cellcolor[RGB]{238,219,159}{58.2} & \cellcolor[RGB]{236,196,160}{58.3} & \nan & \nan & \cellcolor[RGB]{185,214,168}{42.0} & \cellcolor[RGB]{221,223,162}{36.9} & \cellcolor[RGB]{237,227,159}{61.9} & \cellcolor[RGB]{237,209,159}{61.8} & \nan & \nan \\
         & $\mathcal{Q}$ & \cellcolor[RGB]{238,216,159}{33.1} & \cellcolor[RGB]{229,225,161}{36.7} & \cellcolor[RGB]{233,226,160}{33.4} & \cellcolor[RGB]{170,206,163}{87.2} & \cellcolor[RGB]{178,211,166}{85.1} & \cellcolor[RGB]{233,172,165}{7.6} & \cellcolor[RGB]{232,168,168}{0.7} & \cellcolor[RGB]{238,218,159}{33.4} & \cellcolor[RGB]{171,206,163}{45.4} & \cellcolor[RGB]{164,202,160}{88.5} & \cellcolor[RGB]{172,207,164}{87.5} & \cellcolor[RGB]{233,174,164}{9.4} & \cellcolor[RGB]{232,168,168}{0.7} \\
         & $\mathcal{G}_{12}$ & \cellcolor[RGB]{204,219,164}{40.9} & \cellcolor[RGB]{236,227,159}{35.5} & \cellcolor[RGB]{234,227,160}{33.2} & \cellcolor[RGB]{168,205,162}{88.0} & \cellcolor[RGB]{177,210,166}{85.5} & \cellcolor[RGB]{235,179,161}{19.8} & \cellcolor[RGB]{235,179,161}{20.1} & \cellcolor[RGB]{175,209,165}{44.6} & \cellcolor[RGB]{173,208,164}{44.6} & \cellcolor[RGB]{163,202,160}{89.3} & \cellcolor[RGB]{174,208,165}{86.3} & \cellcolor[RGB]{236,188,160}{23.8} & \cellcolor[RGB]{235,187,160}{23.8} \\
         & $\mathcal{H}$ & \cellcolor[RGB]{163,202,160}{51.4} & \cellcolor[RGB]{169,205,162}{48.8} & \cellcolor[RGB]{165,203,161}{55.0} & \cellcolor[RGB]{158,199,158}{95.5} & \cellcolor[RGB]{160,200,159}{95.8} & \cellcolor[RGB]{161,201,159}{87.2} & \cellcolor[RGB]{161,201,159}{92.1} & \cellcolor[RGB]{160,200,159}{49.1} & \cellcolor[RGB]{169,205,162}{46.0} & \cellcolor[RGB]{158,199,158}{92.2} & \cellcolor[RGB]{159,200,159}{94.7} & \cellcolor[RGB]{159,200,158}{81.2} & \cellcolor[RGB]{159,200,159}{83.5} \\
         \hline
         \multicolumn{1}{l}{\parbox[t]{2mm}{\multirow{5}{*}{\rotatebox[origin=c]{90}{\textit{max.$\gamma$}}}}} & $\mathcal{L}$ & \cellcolor[RGB]{211,221,163}{39.8} & \cellcolor[RGB]{215,222,163}{38.9} & \cellcolor[RGB]{207,220,164}{39.6} & \nan & \nan & \nan & \nan & \cellcolor[RGB]{196,217,166}{40.5} & \cellcolor[RGB]{209,220,164}{38.5} & \nan & \nan & \nan & \nan \\
         & $\mathcal{O}$ & \cellcolor[RGB]{213,221,163}{39.5} & \cellcolor[RGB]{235,227,160}{35.8} & \cellcolor[RGB]{236,190,160}{21.0} & \cellcolor[RGB]{193,216,166}{74.9} & \cellcolor[RGB]{237,227,159}{68.7} & \nan & \nan & \cellcolor[RGB]{237,203,160}{31.0} & \cellcolor[RGB]{188,215,167}{41.1} & \cellcolor[RGB]{178,210,166}{79.7} & \cellcolor[RGB]{227,225,161}{70.2} & \nan & \nan \\
         & $\mathcal{Q}$ & \cellcolor[RGB]{238,217,159}{33.4} & \cellcolor[RGB]{237,206,159}{31.0} & \cellcolor[RGB]{209,220,164}{39.0} & \cellcolor[RGB]{170,206,163}{87.2} & \cellcolor[RGB]{179,211,167}{84.6} & \cellcolor[RGB]{233,172,166}{7.1} & \cellcolor[RGB]{233,172,165}{8.2} & \cellcolor[RGB]{238,217,159}{33.2} & \cellcolor[RGB]{208,220,164}{38.6} & \cellcolor[RGB]{166,204,161}{87.2} & \cellcolor[RGB]{172,207,164}{87.0} & \cellcolor[RGB]{233,173,165}{8.2} & \cellcolor[RGB]{233,173,165}{8.4} \\
         & $\mathcal{G}_{12}$ & \cellcolor[RGB]{231,226,160}{36.7} & \cellcolor[RGB]{238,214,159}{32.5} & \cellcolor[RGB]{225,224,161}{35.2} & \cellcolor[RGB]{175,209,165}{83.7} & \cellcolor[RGB]{186,214,167}{81.6} & \cellcolor[RGB]{235,180,161}{20.4} & \cellcolor[RGB]{232,169,167}{2.4} & \cellcolor[RGB]{218,222,162}{37.7} & \cellcolor[RGB]{192,216,166}{40.6} & \cellcolor[RGB]{167,204,162}{86.6} & \cellcolor[RGB]{179,211,167}{83.0} & \cellcolor[RGB]{236,187,160}{23.4} & \cellcolor[RGB]{232,170,167}{2.9} \\
         & $\mathcal{H}$ & \cellcolor[RGB]{158,199,158}{53.3} & \cellcolor[RGB]{158,199,158}{52.7} & \cellcolor[RGB]{173,208,164}{50.4} & \cellcolor[RGB]{161,201,159}{93.6} & \cellcolor[RGB]{162,201,159}{95.0} & \cellcolor[RGB]{167,204,162}{82.1} & \cellcolor[RGB]{171,206,163}{82.3} & \cellcolor[RGB]{158,199,158}{49.7} & \cellcolor[RGB]{161,200,159}{48.4} & \cellcolor[RGB]{160,200,159}{91.0} & \cellcolor[RGB]{160,200,159}{94.4} & \cellcolor[RGB]{161,201,159}{79.9} & \cellcolor[RGB]{166,204,161}{77.2} \\
         \hline
         \multicolumn{1}{l}{\parbox[t]{2mm}{\multirow{5}{*}{\rotatebox[origin=c]{90}{\textit{max.$\alpha$}}}}} & $\mathcal{L}$ & \cellcolor[RGB]{212,221,163}{39.7} & \cellcolor[RGB]{219,223,162}{38.2} & \cellcolor[RGB]{204,219,165}{40.3} & \nan & \nan & \nan & \nan & \cellcolor[RGB]{168,205,162}{46.7} & \cellcolor[RGB]{181,212,167}{42.4} & \nan & \nan & \nan & \nan \\
         & $\mathcal{O}$ & \cellcolor[RGB]{211,221,163}{39.8} & \cellcolor[RGB]{221,223,162}{37.9} & \cellcolor[RGB]{235,184,160}{19.3} & \cellcolor[RGB]{191,215,167}{75.7} & \cellcolor[RGB]{227,225,161}{71.3} & \nan & \nan & \cellcolor[RGB]{233,226,160}{35.7} & \cellcolor[RGB]{238,228,159}{34.7} & \cellcolor[RGB]{177,210,166}{80.3} & \cellcolor[RGB]{213,221,163}{73.7} & \nan & \nan \\
         & $\mathcal{Q}$ & \cellcolor[RGB]{239,224,159}{34.8} & \cellcolor[RGB]{226,225,161}{37.1} & \cellcolor[RGB]{219,223,162}{36.6} & \cellcolor[RGB]{169,206,163}{87.5} & \cellcolor[RGB]{179,211,167}{84.7} & \cellcolor[RGB]{233,172,165}{7.9} & \cellcolor[RGB]{233,171,166}{6.4} & \cellcolor[RGB]{181,213,168}{42.6} & \cellcolor[RGB]{181,212,168}{42.2} & \cellcolor[RGB]{164,202,160}{88.4} & \cellcolor[RGB]{173,208,164}{86.7} & \cellcolor[RGB]{234,175,164}{10.8} & \cellcolor[RGB]{233,172,165}{6.8} \\
         & $\mathcal{G}_{12}$ & \cellcolor[RGB]{234,227,160}{36.3} & \cellcolor[RGB]{226,224,161}{37.2} & \cellcolor[RGB]{234,227,160}{33.2} & \cellcolor[RGB]{171,207,163}{86.3} & \cellcolor[RGB]{179,211,167}{84.6} & \cellcolor[RGB]{234,176,163}{14.2} & \cellcolor[RGB]{234,177,162}{17.3} & \cellcolor[RGB]{194,216,166}{40.8} & \cellcolor[RGB]{163,202,160}{47.7} & \cellcolor[RGB]{162,202,160}{89.4} & \cellcolor[RGB]{174,208,165}{86.1} & \cellcolor[RGB]{234,179,161}{17.1} & \cellcolor[RGB]{235,181,160}{20.7} \\
         & $\mathcal{H}$ & \cellcolor[RGB]{161,201,159}{52.3} & \cellcolor[RGB]{160,200,159}{51.9} & \cellcolor[RGB]{166,203,161}{54.6} & \cellcolor[RGB]{160,200,159}{93.8} & \cellcolor[RGB]{161,201,159}{95.1} & \cellcolor[RGB]{167,204,162}{82.2} & \cellcolor[RGB]{170,206,163}{83.2} & \cellcolor[RGB]{167,204,162}{47.0} & \cellcolor[RGB]{158,199,158}{49.2} & \cellcolor[RGB]{159,200,159}{91.4} & \cellcolor[RGB]{159,200,159}{94.7} & \cellcolor[RGB]{161,201,159}{79.8} & \cellcolor[RGB]{165,203,161}{78.0} \\
         \hline
         \multicolumn{1}{l}{\parbox[t]{2mm}{\multirow{5}{*}{\rotatebox[origin=c]{90}{\textit{max.$\sigma$}}}}} & $\mathcal{L}$ & \cellcolor[RGB]{205,219,164}{40.8} & \cellcolor[RGB]{239,227,159}{34.9} & \cellcolor[RGB]{232,168,168}{4.9} & \nan & \nan & \nan & \nan & \cellcolor[RGB]{174,208,165}{45.0} & \cellcolor[RGB]{179,211,167}{42.9} & \nan & \nan & \nan & \nan \\
         & $\mathcal{O}$ & \cellcolor[RGB]{231,226,160}{36.7} & \cellcolor[RGB]{232,226,160}{36.2} & \cellcolor[RGB]{236,189,160}{20.7} & \cellcolor[RGB]{197,217,166}{73.9} & \cellcolor[RGB]{238,219,159}{65.4} & \nan & \nan & \cellcolor[RGB]{199,217,165}{40.2} & \cellcolor[RGB]{214,221,163}{37.9} & \cellcolor[RGB]{180,212,167}{77.9} & \cellcolor[RGB]{239,223,159}{65.8} & \nan & \nan \\
         & $\mathcal{Q}$ & \cellcolor[RGB]{223,224,162}{38.0} & \cellcolor[RGB]{226,225,161}{37.1} & \cellcolor[RGB]{201,218,165}{40.9} & \cellcolor[RGB]{168,205,162}{88.1} & \cellcolor[RGB]{177,210,166}{85.7} & \cellcolor[RGB]{232,169,167}{2.3} & \cellcolor[RGB]{232,169,168}{1.1} & \cellcolor[RGB]{176,209,165}{44.4} & \cellcolor[RGB]{201,218,165}{39.4} & \cellcolor[RGB]{163,202,160}{88.7} & \cellcolor[RGB]{171,206,163}{88.0} & \cellcolor[RGB]{232,170,167}{2.9} & \cellcolor[RGB]{232,168,168}{0.7} \\
         & $\mathcal{G}_{12}$ & \cellcolor[RGB]{206,219,164}{40.6} & \cellcolor[RGB]{214,221,163}{39.0} & \cellcolor[RGB]{231,226,160}{33.8} & \cellcolor[RGB]{170,206,163}{87.0} & \cellcolor[RGB]{174,208,164}{87.8} & \cellcolor[RGB]{234,177,162}{15.9} & \cellcolor[RGB]{232,169,168}{1.1} & \cellcolor[RGB]{173,208,164}{45.3} & \cellcolor[RGB]{180,212,167}{42.6} & \cellcolor[RGB]{163,202,160}{89.0} & \cellcolor[RGB]{169,205,163}{89.1} & \cellcolor[RGB]{235,181,160}{19.9} & \cellcolor[RGB]{232,169,168}{1.3} \\
         & $\mathcal{H}$ & \cellcolor[RGB]{180,212,167}{45.0} & \cellcolor[RGB]{234,176,163}{23.1} & \cellcolor[RGB]{166,204,161}{54.3} & \cellcolor[RGB]{160,200,159}{94.1} & \cellcolor[RGB]{164,202,160}{93.7} & \cellcolor[RGB]{159,199,158}{89.7} & \cellcolor[RGB]{159,199,158}{94.3} & \cellcolor[RGB]{165,203,161}{47.5} & \cellcolor[RGB]{232,168,168}{20.0} & \cellcolor[RGB]{160,200,159}{91.2} & \cellcolor[RGB]{163,202,160}{92.8} & \cellcolor[RGB]{158,199,158}{81.9} & \cellcolor[RGB]{158,199,158}{84.4} \\
         \hline
         \multicolumn{1}{l}{\parbox[t]{2mm}{\multirow{5}{*}{\rotatebox[origin=c]{90}{\textit{mid.$\alpha$}}}}} & $\mathcal{L}$ & \cellcolor[RGB]{210,220,164}{39.9} & \cellcolor[RGB]{206,219,164}{40.2} & \cellcolor[RGB]{209,220,164}{39.1} & \nan & \nan & \nan & \nan & \cellcolor[RGB]{160,200,159}{49.3} & \cellcolor[RGB]{215,222,163}{37.7} & \nan & \nan & \nan & \nan \\
         & $\mathcal{O}$ & \cellcolor[RGB]{234,176,163}{23.1} & \cellcolor[RGB]{226,225,161}{37.1} & \cellcolor[RGB]{237,204,160}{25.1} & \cellcolor[RGB]{235,182,160}{44.5} & \cellcolor[RGB]{233,171,166}{42.7} & \nan & \nan & \cellcolor[RGB]{232,168,168}{20.0} & \cellcolor[RGB]{213,221,163}{37.9} & \cellcolor[RGB]{235,185,160}{47.4} & \cellcolor[RGB]{233,172,165}{44.0} & \nan & \nan \\
         & $\mathcal{Q}$ & \cellcolor[RGB]{239,227,159}{35.2} & \cellcolor[RGB]{238,215,159}{32.8} & \cellcolor[RGB]{228,225,161}{34.5} & \cellcolor[RGB]{171,206,163}{86.3} & \cellcolor[RGB]{178,211,167}{84.8} & \cellcolor[RGB]{233,171,166}{6.1} & \cellcolor[RGB]{233,171,166}{5.6} & \cellcolor[RGB]{191,215,167}{41.2} & \cellcolor[RGB]{235,180,160}{26.9} & \cellcolor[RGB]{164,203,161}{88.2} & \cellcolor[RGB]{174,208,164}{86.4} & \cellcolor[RGB]{233,173,165}{7.3} & \cellcolor[RGB]{233,171,166}{5.6} \\
         & $\mathcal{G}_{12}$ & \cellcolor[RGB]{218,223,162}{38.7} & \cellcolor[RGB]{238,214,159}{32.5} & \cellcolor[RGB]{237,209,159}{26.6} & \cellcolor[RGB]{170,206,163}{86.6} & \cellcolor[RGB]{175,209,165}{86.8} & \cellcolor[RGB]{232,168,168}{0.5} & \cellcolor[RGB]{232,169,168}{1.1} & \cellcolor[RGB]{174,208,165}{44.8} & \cellcolor[RGB]{169,205,163}{45.8} & \cellcolor[RGB]{164,202,160}{88.5} & \cellcolor[RGB]{170,206,163}{88.4} & \cellcolor[RGB]{232,168,168}{0.5} & \cellcolor[RGB]{232,169,168}{1.0} \\
         & $\mathcal{H}$ & \cellcolor[RGB]{193,216,166}{42.6} & \cellcolor[RGB]{232,168,168}{17.6} & \cellcolor[RGB]{170,206,163}{52.4} & \cellcolor[RGB]{161,201,159}{93.5} & \cellcolor[RGB]{161,201,159}{95.2} & \cellcolor[RGB]{159,199,158}{89.6} & \cellcolor[RGB]{159,199,158}{94.1} & \cellcolor[RGB]{158,199,158}{49.8} & \cellcolor[RGB]{232,168,168}{20.0} & \cellcolor[RGB]{160,200,159}{90.9} & \cellcolor[RGB]{161,201,159}{93.6} & \cellcolor[RGB]{159,200,158}{81.3} & \cellcolor[RGB]{158,199,158}{84.2} \\
         \hline
         \multicolumn{1}{l}{\parbox[t]{2mm}{\multirow{5}{*}{\rotatebox[origin=c]{90}{\textit{min.$\gamma$}}}}} & $\mathcal{L}$ & \cellcolor[RGB]{239,226,159}{35.1} & \cellcolor[RGB]{234,176,163}{23.1} & \cellcolor[RGB]{232,168,168}{4.9} & \nan & \nan & \nan & \nan & \cellcolor[RGB]{192,216,166}{41.1} & \cellcolor[RGB]{232,168,168}{20.0} & \nan & \nan & \nan & \nan \\
         & $\mathcal{O}$ & \cellcolor[RGB]{238,228,159}{35.7} & \cellcolor[RGB]{210,220,164}{39.6} & \cellcolor[RGB]{237,199,160}{23.5} & \cellcolor[RGB]{233,171,166}{30.6} & \cellcolor[RGB]{233,173,165}{44.8} & \nan & \nan & \cellcolor[RGB]{208,220,164}{39.0} & \cellcolor[RGB]{202,218,165}{39.3} & \cellcolor[RGB]{233,170,167}{33.1} & \cellcolor[RGB]{233,173,165}{44.5} & \nan & \nan \\
         & $\mathcal{Q}$ & \cellcolor[RGB]{192,216,166}{42.7} & \cellcolor[RGB]{212,221,163}{39.3} & \cellcolor[RGB]{239,222,159}{30.3} & \cellcolor[RGB]{171,206,163}{86.4} & \cellcolor[RGB]{177,210,166}{85.5} & \cellcolor[RGB]{232,168,168}{1.0} & \cellcolor[RGB]{232,168,168}{0.3} & \cellcolor[RGB]{236,227,159}{35.3} & \cellcolor[RGB]{234,177,163}{24.9} & \cellcolor[RGB]{165,203,161}{87.8} & \cellcolor[RGB]{172,207,164}{87.4} & \cellcolor[RGB]{232,169,168}{1.1} & \cellcolor[RGB]{232,168,168}{0.3} \\
         & $\mathcal{G}_{12}$ & \cellcolor[RGB]{203,218,165}{41.1} & \cellcolor[RGB]{197,217,166}{41.5} & \cellcolor[RGB]{238,211,159}{27.1} & \cellcolor[RGB]{171,207,163}{86.3} & \cellcolor[RGB]{174,208,165}{87.5} & \cellcolor[RGB]{232,168,168}{0.7} & \cellcolor[RGB]{232,168,168}{0.7} & \cellcolor[RGB]{176,210,166}{44.2} & \cellcolor[RGB]{215,222,163}{37.6} & \cellcolor[RGB]{164,202,160}{88.4} & \cellcolor[RGB]{170,206,163}{88.7} & \cellcolor[RGB]{232,168,168}{0.6} & \cellcolor[RGB]{232,168,168}{0.6} \\
         & $\mathcal{H}$ & \cellcolor[RGB]{232,168,168}{17.6} & \cellcolor[RGB]{236,199,160}{29.6} & \cellcolor[RGB]{190,215,167}{43.5} & \cellcolor[RGB]{160,200,159}{93.9} & \cellcolor[RGB]{167,204,162}{91.5} & \cellcolor[RGB]{159,199,158}{89.9} & \cellcolor[RGB]{158,199,158}{94.6} & \cellcolor[RGB]{232,168,168}{20.0} & \cellcolor[RGB]{232,226,160}{35.5} & \cellcolor[RGB]{160,200,159}{91.2} & \cellcolor[RGB]{166,204,161}{90.8} & \cellcolor[RGB]{159,199,158}{81.5} & \cellcolor[RGB]{158,199,158}{84.3} \\
         \hline
         \multicolumn{1}{l}{\parbox[t]{2mm}{\multirow{5}{*}{\rotatebox[origin=c]{90}{\textit{min.$\alpha$}}}}} & $\mathcal{L}$ & \cellcolor[RGB]{236,227,160}{35.9} & \cellcolor[RGB]{237,228,159}{35.4} & \cellcolor[RGB]{233,171,166}{7.5} & \nan & \nan & \nan & \nan & \cellcolor[RGB]{188,215,167}{41.6} & \cellcolor[RGB]{239,223,159}{33.8} & \nan & \nan & \nan & \nan \\
         & $\mathcal{O}$ & \cellcolor[RGB]{232,169,168}{18.1} & \cellcolor[RGB]{215,222,163}{38.9} & \cellcolor[RGB]{236,191,160}{21.2} & \cellcolor[RGB]{232,168,168}{27.2} & \cellcolor[RGB]{232,168,168}{39.2} & \nan & \nan & \cellcolor[RGB]{232,168,168}{20.0} & \cellcolor[RGB]{169,206,163}{45.8} & \cellcolor[RGB]{232,168,168}{30.5} & \cellcolor[RGB]{232,168,168}{39.2} & \nan & \nan \\
         & $\mathcal{Q}$ & \cellcolor[RGB]{237,209,159}{31.8} & \cellcolor[RGB]{237,227,159}{35.4} & \cellcolor[RGB]{238,217,159}{28.8} & \cellcolor[RGB]{172,207,164}{85.6} & \cellcolor[RGB]{178,211,166}{85.1} & \cellcolor[RGB]{232,168,168}{0.2} & \cellcolor[RGB]{232,168,168}{0.0} & \cellcolor[RGB]{211,220,163}{38.6} & \cellcolor[RGB]{197,217,166}{40.0} & \cellcolor[RGB]{165,203,161}{87.9} & \cellcolor[RGB]{172,207,164}{87.2} & \cellcolor[RGB]{232,168,168}{0.2} & \cellcolor[RGB]{232,168,168}{0.0} \\
         & $\mathcal{G}_{12}$ & \cellcolor[RGB]{215,222,163}{39.1} & \cellcolor[RGB]{238,213,159}{32.4} & \cellcolor[RGB]{237,207,159}{25.9} & \cellcolor[RGB]{173,207,164}{85.1} & \cellcolor[RGB]{175,209,165}{87.0} & \cellcolor[RGB]{232,168,168}{0.2} & \cellcolor[RGB]{232,168,168}{0.0} & \cellcolor[RGB]{189,215,167}{41.5} & \cellcolor[RGB]{213,221,163}{38.0} & \cellcolor[RGB]{164,203,161}{88.1} & \cellcolor[RGB]{170,206,163}{88.6} & \cellcolor[RGB]{232,168,168}{0.3} & \cellcolor[RGB]{232,168,168}{0.0} \\
         & $\mathcal{H}$ & \cellcolor[RGB]{232,168,168}{17.6} & \cellcolor[RGB]{237,200,160}{30.0} & \cellcolor[RGB]{232,168,168}{4.9} & \cellcolor[RGB]{160,200,159}{93.8} & \cellcolor[RGB]{160,200,159}{95.8} & \cellcolor[RGB]{159,199,158}{89.6} & \cellcolor[RGB]{159,200,158}{93.9} & \cellcolor[RGB]{232,168,168}{20.0} & \cellcolor[RGB]{219,223,162}{37.1} & \cellcolor[RGB]{160,200,159}{90.7} & \cellcolor[RGB]{160,200,159}{94.4} & \cellcolor[RGB]{159,200,158}{81.3} & \cellcolor[RGB]{158,199,158}{84.4} \\
         \hline
    \end{tabular}
    \caption{\label{fig:subset-multilingual}Performance comparison on sample subsets for Spanish and French datasets. Same criteria and notation as in Table \ref{fig:subsets_roberta}. We use ISO-639 codes to denote languages.}
\end{table*}

\end{document}